\pdfoutput=1

\documentclass[11pt]{article}

\usepackage[]{acl}

\usepackage{times}
\usepackage{latexsym}

\usepackage[T1]{fontenc}

\usepackage[utf8]{inputenc}

\usepackage{microtype}

\usepackage{enumitem}

\usepackage{graphicx}
\usepackage{xcolor,colortbl}
\usepackage{amsmath}
\usepackage{booktabs}
\usepackage{multirow}
\usepackage{array}
\usepackage{comment}
\usepackage{bm}
\usepackage{multirow}
\usepackage{scalerel}
\usepackage{textcomp}
\usepackage{microtype}
\usepackage{amssymb}
\usepackage{pifont}
\newcommand{\cmark}{\ding{51}}%
\newcommand{\xmark}{\ding{55}}%

\usepackage{array}
\newcolumntype{P}[1]{>{\centering\arraybackslash}p{#1}}
\newcolumntype{M}[1]{>{\centering\arraybackslash}m{#1}}

\newcommand\given[1][]{\:#1\vert\:}
\newcommand{\STAB}[1]{\begin{tabular}{@{}c@{}}#1\end{tabular}}
\newcommand\redtable[1]{\underline{#1}}

\definecolor{nicegreen}{HTML}{009B55}
\definecolor{niceorange}{HTML}{F86624}
\definecolor{niceblue}{HTML}{072AC8}
\newcommand{\gr}[1]{\textcolor{niceblue}{#1}}
\newcommand\red[1]{\textcolor{niceorange}{#1}}
\def\dancer{\scalerel*{\includegraphics{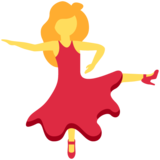}}{\textrm{\textbigcircle}}}

\newcommand{\dataset}{VALSE} 

\title{\dataset{} \dancer: A Task-Independent Benchmark for Vision and Language Models Centered on Linguistic Phenomena}
\author{Letitia Parcalabescu$^1$\thanks{~~Corresponding author \url{parcalabescu@cl.uni-heidelberg.de}.}\ \ ~~Michele Cafagna$^2$ \ \ ~~Lilitta Muradjan$^1$ \\ ~~\textbf{Anette Frank}$^1$\ \ ~~\textbf{Iacer Calixto}$^{3,4}$\ \ ~~\textbf{Albert Gatt}$^{5,2}$ \\\\
$^1$Heidelberg University, Department of Computational Linguistics\\ 
$^2$University of Malta, Institute of Linguistics and Language Technology \\
$^3$New York University \ \ 
$^4$ILLC, University of Amsterdam\\
$^5$Utrecht University, Department of Information and Computing Sciences}

\date{}

\begin{document}
\maketitle

\begin{abstract}
We propose \dataset{} (\textbf{V}ision \textbf{A}nd \textbf{L}anguage \textbf{S}tructured \textbf{E}valuation), a novel benchmark designed for 
testing general-purpose pretrained vision and language (V\&L) models for their {\em visio-linguistic grounding} capabilities on {\em specific linguistic phenomena}. 
\dataset{} 
offers a suite of six tests covering various 
linguistic constructs. Solving these 
requires models to ground linguistic phenomena in the visual modality, allowing more fine-grained evaluations than hitherto possible.
We build \dataset{} using methods that support the construction of 
\textit{valid}
foils, and report results from evaluating five widely-used V\&L models.
Our experiments suggest that current models have considerable difficulty addressing most phenomena.
Hence, we expect \dataset{} to serve as an important benchmark to measure future progress of pretrained V\&L models from a \textit{linguistic perspective}, complementing the canonical task-centred V\&L evaluations.
\end{abstract}

\section{Introduction}\label{sec:intro}
General-purpose pretrained vision and language (V\&L) models have gained notable performance on many V\&L tasks
\cite{lu2019vilbert,tan-bansal-2019-lxmert,li2019visualbert,chen2020uniter,Li-etal-2020unicodervl,Su2020VL-BERT}.
As a result, 
V\&L research has changed its focus
from task-specific
architectures 
to fine-tuning
large
V\&L models.

Current benchmarks give a good perspective on 
model performance on a wide range of 
V\&L tasks \citep{cao2020behind,lourie2021unicorn,li2021value}, but the field is only starting to assess \emph{why} models perform so well 
and whether models learn \emph{specific capabilities that span multiple V\&L tasks}. 
Specifically, we lack detailed understanding of the extent to which such models are able to ground 
linguistic phenomena---from
morphosyntax 
to semantics---in the visual modality \cite{Bernardi2021}. For example, recent evidence suggests that models are insensitive to linguistic distinctions 
of verb-argument structure \cite{Hendricks2021} and word order \cite{cirik-etal-2018-visual, akula-etal-2020-words}.



Our work addresses this gap with \dataset{} (Vision And Language Structured Evaluation), a benchmark for V\&L model evaluation 
comprising six 
tasks, or `pieces', where each piece has the same structure: given a visual input, a 
model is 
asked to distinguish real captions from \textit{foils},
where a foil is constructed from a caption by altering a word or phrase that realizes
a \emph{specific linguistic phenomenon}, e.g.,
semantic number 
of nouns, verb argument structure, or coreference. 
\dataset{} uses a resource-lean diagnostic setup that  
dispenses with large-scale annotation 
(e.g., of bounding boxes), and builds on existing high-quality image captioning and VQA data.
\dataset{} is designed to leverage the existing prediction heads 
in pretrained (or finetuned) V\&L models; for that reason, our benchmark does not include any re-training and can be interpreted as a \textit{zero-shot} evaluation.
We build \textit{test} data for each piece so as to safeguard against the possibility of models exploiting artefacts or statistical biases in the data, a well-known issue with highly parameterised neural models pretrained on large amounts of data \citep{goyal2017making,madhyastha-etal-2018-defoiling,10.3389/frai.2019.00028}.
With this in view, we propose novel methods to guard against the emergence of \emph{artefacts} during foiling.


Our main contributions are:

\begin{enumerate}[label=\roman*), noitemsep]
    \item We introduce \dataset{}, a novel benchmark aimed at gauging the sensitivity of 
    pre-trained V\&L models 
    to \emph{foiled} instances.
    \item We cover a wide spectrum of basic linguistic phenomena affecting the linguistic \emph{and} visual modalities: existence, plurality, counting, spatial relations, actions, and entity coreference.   
    \item We investigate 
    novel strategies to build \textit{valid}
    foils that include automatic \textit{and} human validation.
    We balance 
    \textit{word frequency distributions} between captions and foils, and test against 
    pretrained models 
    solving the benchmark \textit{unimodally} by relying only on text. 
    We employ
    \textit{masked language modeling} (MLM)
    in foil creation and \textit{semantic inference} 
    for validating foils, and 
    finally collect \textit{human annotations} for the entire benchmark.
    \item We establish initial experimental results for
    pretrained V\&L models 
    of diverse architectures on \dataset{}.
   The overall weak performance of these models
    indicates that the time is ripe for a 
    novel, reliable foiling dataset targeting
    the 
    visual grounding capabilities of V\&L models through the lens of linguistic constructs.\footnote{We release our dataset containing all annotators' votes \cite{prabhakaran2021releasing} at \url{https://github.com/Heidelberg-NLP/VALSE}. }
\end{enumerate}

\section{Background and Related work}\label{sec:related_work}

Pretrained V\&L models learn to combine vision and language through self-supervised multitask learning. Tasks include \textit{multimodal masked modeling}---where words in the text and object labels or regions in the image are masked out, then predicted---and \textit{image-sentence alignment}, whereby a model learns to predict whether an image and a text correspond. 
Major architectures are single- and dual-stream multimodal transformers: \textit{single-stream models}
concatenate word and image features, and encode the resulting sequence with a single transformer stack;
\textit{dual-stream models}
use distinct transformer stacks to handle visual and textual inputs, and additional layers (e.g. co-attention) to fuse these into multimodal features.

\begin{table*}[t]
    \small
    \centering
    
    \resizebox{\linewidth}{!}{
    \begin{tabular}{%
    p{.01\linewidth}%
    >{\raggedleft\arraybackslash}p{.1\linewidth}%
    p{.09\linewidth}%
    p{.16\linewidth}%
    p{.17\linewidth}p{.13\linewidth}p{.12\linewidth}p{.12\linewidth}}
      \toprule
      \multirow{13}{*}{\STAB{\rotatebox[origin=c]{90}{\bf Data collection \& metadata}}} 
      & pieces & \bf existence & \bf plurality & \bf counting & \bf relations & \bf actions & \bf coreference \\
      \cmidrule{3-8}
      & instruments & \emph{existential quantifiers} & \emph{semantic number} & \emph{balanced}, \emph{adver\-sarial}, \emph{small numbers} & \emph{prepositions} & \emph{replacement}, \emph{actant swap} & \emph{standard}, \emph{clean} \\
      & \#examples$^\dagger$ & $505$ & $851$ & $2,459$ & $535$ & $1,633$ & $812$ \\
      \cmidrule{3-8}
      & foil generation method & \emph{nothing} $\leftrightarrow$ \emph{something} &  NP replacement ({\tt sg2pl}; {\tt pl2sg}) \& quantifier insertion & numeral ~~~~~~~~~~~~~~~~~~~ replacement & SpanBERT prediction & action replacement, actant swap  & \emph{yes} $\leftrightarrow$ \emph{no}\\
      \cmidrule{3-8}
      & MLM   & \xmark{} & \xmark{}  & \xmark{} & \cmark{} & \cmark{} & \xmark{} \\
      & GRUEN   & \xmark{} &  \cmark{}   &   \xmark{}   &  \cmark{} & \xmark{} & \xmark{}\\
      & NLI   & \xmark{} &  \cmark{}   &   \xmark{}   &  \cmark{} & \xmark{} & \xmark{}\\
      & src. dataset & Visual7W & MSCOCO & Visual7W & MSCOCO & SWiG  & VisDial v1.0\\
      & image src. & MSCOCO & MSCOCO & MSCOCO & MSCOCO & SituNet & MSCOCO\\
      \midrule
      \multirow{10}{*}{\STAB{\rotatebox[origin=c]{90}{\bf Example data}}}
      & caption (\gr{blue}) / foil (\red{orange}) &
      \textit{There are \gr{no animals} / \red{animals} shown.}&
      \textit{A small copper vase with \gr{some flowers} / \red{exactly one flower} in it.
      } &
      \textit{There are \gr{four} / \red{six} zebras.} &
      \textit{A cat plays with a pocket knife \gr{on} / \red{underneath} a table.} &
      \textit{A \gr{man} / \red{woman} shouts at a \gr{woman} / \red{man}.} &
      \textit{Buffalos walk along grass. Are they in a zoo? \gr{No} / \red{Yes}.}\\
      \cmidrule{3-8}
      & image             &
      \includegraphics[width=\linewidth]{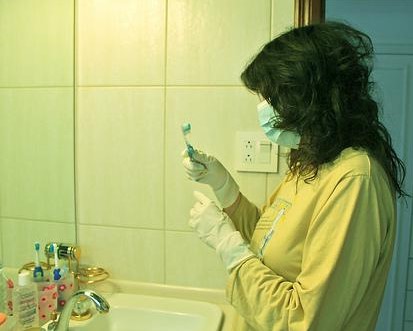}&
      \centering \includegraphics[width=0.58\linewidth]{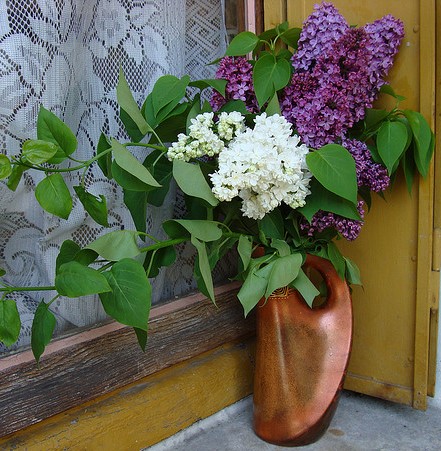} &
      \includegraphics[width=0.79\linewidth]{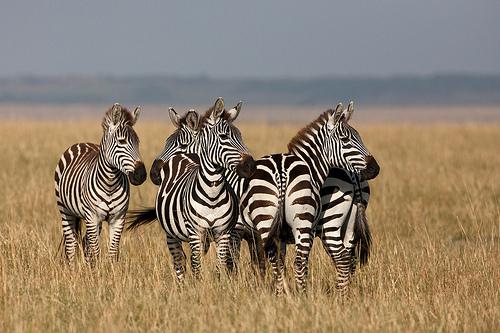} & 
      \includegraphics[width=.8\linewidth]{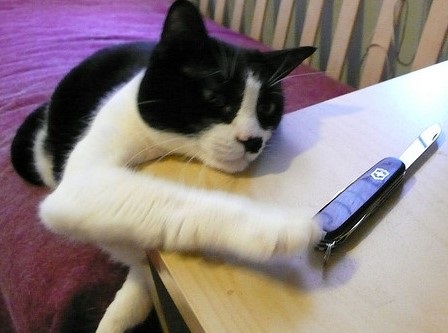} &
      \includegraphics[width=\linewidth]{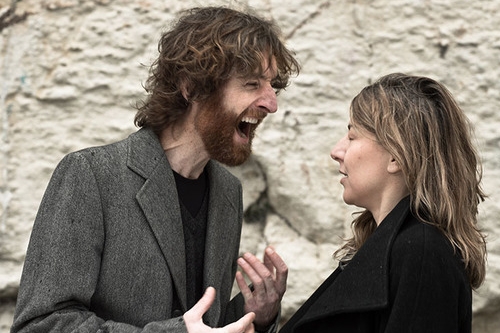} &
      \includegraphics[width=\linewidth]{imgs/coref_example.jpg} \\
      \bottomrule
    \end{tabular}
    }
    \caption{Overview of pieces and instruments in  \dataset{}, with number of examples per piece; the foil generation method used; whether masked language modelling (MLM), GRUEN, and NLI filtering are used; dataset and image sources; and image-caption-foil examples.
    $^\dagger$The number of examples is the sum of the examples available for each instrument in the piece.
    In Table \ref{tab:mturk-results} (in the Appendix) we list the number of examples in each individual instrument.}
    \label{tab:phenomena}
\end{table*}

\paragraph{Benchmarking V\&L models}
V\&L models \cite{li2019visualbert,lu2019vilbert,tan-bansal-2019-lxmert,lu2020vilbert12in1, li2020oscar, kim2021vilt}
are commonly evaluated on V\&L \textit{tasks} such as VQA \citep{goyal2017making}, visual reasoning \citep{suhr2018corpus}, or image retrieval \citep{Lin-etal:2014:mscoco,plummer2015flickr30k}.

Given how well transformer-based models perform across unimodal and multimodal tasks, research efforts have recently started to address what makes them so effective, and to what extent they learn generalisable representations. Techniques to address these questions 
in unimodal and multimodal V\&L contexts 
include:
adversarial examples \citep{jia-liang-2017-adversarial,jia-etal-2019-certified}; investigation of the impact of bias, 
be it
linguistic \citep{gururangan-etal-2018-annotation}, visual semantic \citep{agarwal2020towards}, or socio-economic \citep{garg-etal-2019-counterfactual}; and the use of 
linguistically-informed counterfactual and minimally-edited examples \citep{levesque2012winograd,gardner-etal-2020-evaluating}.
A trend within the latter research line that is specific to V\&L models is \textit{vision-and-language foiling} \citep{shekhar-etal-2017-foil, gokhale-etal-2020-mutant,bitton-etal-2021-automatic,parcalabescu2021seeing,rosenberg-etal-2021-vqa}, where the idea is to create counterfactual (i.e., \textit{foiled}) and/or minimally edited examples by performing data augmentation on captions \citep{shekhar-etal-2017-foil,shekhar-etal-2017-vision} or images \citep{rosenberg-etal-2021-vqa}. 


Since most V\&L models are pretrained on some version of the image-text alignment task, it is possible to test their ability to distinguish correct from foiled captions (in relation to an image) in a zero-shot setting. The construction of foils can serve many investigation purposes. With \dataset{}, we target the \textit{linguistic grounding capabilities of V\&L models}, focusing
on pervasive linguistic phenomena that span \textit{multiple tokens}, described in \S 3.1--\S3.6.
At the same time, we ensure that our data is robust to 
perturbations and artefacts by
i) controlling for word frequency biases between captions and foils, and ii) testing against \emph{unimodal collapse}, a known
issue of
V\&L models \cite{goyal2017making,madhyastha-etal-2018-defoiling}, thereby preventing models from solving the task using a single input modality.
The issue of neural models exploiting data artefacts is well-known \citep{gururangan-etal-2018-annotation,jia-etal-2019-certified,wang-etal-2020-gradient,HE2021104194} and methods have been proposed to uncover such effects,
including gradient-based, adversarial perturbations or input reduction techniques 
(cf.\ \citealp{wallace-etal-2020-interpreting}).
Yet, these methods are still not fully understood \cite{HE2021104194} and can be unreliable \citep{wang-etal-2020-gradient}.

Our work is related to \citet{gardner-etal-2020-evaluating}, who construct \textit{task-specific} \textit{contrast sets} for NLU.
However, our focus is on modelling \textit{linguistic phenomena} instead of tasks, and we construct carefully curated, balanced, single foils from valid instances that we select from multiple multimodal datasets.

\section{Constructing the \dataset{} benchmark} \label{sec:construct_bench}

We resort to a musical analogy to describe \dataset{}: Vision And Language Structured Evaluation is composed of 6 \emph{pieces}, each corresponding to a specific linguistic phenomenon (see Table~\ref{tab:phenomena} for an overview). Each piece consists of one or more \emph{instruments} designed to evaluate a model's ability to ground that specific linguistic phenomenon.

All instruments are built by applying \textit{foiling functions} (FFs) specific to the linguistic phenomenon under study. FFs take a \textit{correct caption} as input and change a specific part 
to produce a \textit{foiled caption} (or \textit{foil}). 
We design FFs such that the sentences they produce fail to describe the image, while still being grammatical and otherwise valid sentences. 

Of course, a \textit{foiled} caption may be
less likely than the original caption from which it was produced, and such
unwarranted biases 
can be easily picked up
by overparameterised V\&L models.
Moreover, an automatic FF
may fail to produce a foil that contradicts the image, for example by altering the original caption to yield a near-synonymous one, or one that is entailed by the original caption.
For phenomena that make it difficult to control these crucial properties of foils, we apply additional filters:
i) some FFs make use of strong LMs to propose changes to captions, so that the generated foils are still high-probability sentences; ii) we use state-of-the-art natural language inference (NLI) methods
to detect cases where there is an \emph{entailment} between caption and foil, and filter out such foils from the dataset (see \S 4 for discussion).
As a final measure, we employ human annotators to validate all generated testing data in \dataset{}.

\dataset{} data is sourced from
existing V\&L datasets.
Below, we describe each piece and its instruments, and the corresponding task setup in \dataset{}.  For each instrument, we follow the same procedure:
i) we identify captions that contain instances of the targeted linguistic phenomenon;
ii) we apply a FF that automatically replaces the expression with a variant that contradicts the original expression's visual content, thereby constructing one or more foils from each target instance in the original caption, as discussed in \S 4; we then
iii) subject the obtained foils to various filters, with the aim of distilling a subset of \textit{valid} and \textit{reliable} foils that cannot be easily tricked by a new generation of highly parameterised pretrained V\&L models.



\subsection{Existence}\label{subsec:existence}
The {\bf existence} piece has a single instrument and targets instances with \textbf{existential quantifiers}.
Models need to differentiate between examples i) where \textit{there is no entity} of a certain type or ii) where \textit{one or more of these entities} are visible in an image.

We use the Visual7W visual question answering dataset \citep{zhu2016cvpr} and source its `how many' examples, building a pool of those whose answers are numerals (0, 1, 2, etc.).
We use
templates
to transform question and answer fields into a declarative statement that correctly describes what can be seen in the image, e.g.
`Q: How many animals are shown? A: 0' $\rightarrow$ `There are 0 animals shown'.
We then transform these statements into an existential statement. In the example above, we replace the numeral by the word `no' to create a correct caption
(`There are no animals shown')
and remove the numeral altogether to create a foil
(`There are animals shown').
The existence piece has
505 image--caption--foil tuples after manual validation, out of 534 candidates (cf. \S \ref{sec:foiling}),
and captions/foils are balanced: 50\% of the (correct) captions originally have answer 0, and the remaining have answer 1 or greater.
Full details are provided in \ref{app:existence}.


\subsection{Plurality}
The \textbf{plurality} piece has a single instrument, concerned with {\bf semantic number}. It is intended to test whether a model is able to distinguish between noun phrases denoting a single entity in an image (`exactly one flower'), versus multiple entities (`some flowers'). The dataset consists of
851 validated instances out of 1000 generated candidates (cf. \S \ref{sec:foiling}),
evenly divided between cases where the caption contains a plural NP, foiled by replacing it with a singular ({\tt pl2sg}: `some flowers' $\rightarrow$ `exactly one flower'), or conversely, the caption contains a singular which is foiled by replacing it with a plural ({\tt sg2pl}). 
Foil candidates were generated from the COCO 2017 validation set \cite{Chen2015}.
Full details about the foil construction and our measures against introducing biases with quantifiers such as `exactly one', are provided in \ref{app:plurality}.

\subsection{Counting}\label{subsec:counting}
The {\bf counting} piece has three instruments: {\bf balanced}, {\bf adversarial} and {\bf small numbers}.
All instances are \textit{statements about the number of entities visible in an image}.
The model needs to differentiate between examples where \textit{the specific number of entities in the associated image} is correct or incorrect, given the statement.
Similarly to the existence piece, we use the Visual7W VQA dataset \citep{zhu2016cvpr} and source its `how many' examples whose answers are numerals (0, 1, 2, etc.).
We use 
templates
to transform question and answer fields into a declarative statement describing the image and create foils by replacing the numeral in the correct statement by another numeral.


All three instruments are designed to show whether models learn strategies that generalize beyond the training distribution, and to what extent a model exploits class frequency bias.\footnote{We take the original answer in Visual7W as the example class: e.g., in `There are 0 animals shown', the class is 0.}
In {\bf counting balanced} we cap the number of examples to a maximum per class and make sure correct and foil classes are balanced, so that models that exploit class frequency bias are penalized.
In {\bf counting adversarial} we ensure that all foils take class $n \in \{0,1,2,3\}$, whereas all correct captions take class $m \in \{m \given m \geq 4\}$. Biased models are expected to favour more frequent classes.
Since small numbers are naturally the most frequent, models that resort to such biases should perform poorly on this 
adversarial test set.
{\bf Counting small numbers} is a sanity check where all correct captions and foils have class $n \in \{0,1,2,3\}$, and caption/foil classes are balanced. Since models likely have been exposed to many examples in this class set and all such classes are high-frequency, with this instrument we disentangle model performance from class exposure.
Counting balanced, adversarial, and small numbers have 868 (1000), 691 (756), and 900 (1000) instances after (before) manual validation, respectively (cf. \S \ref{sec:foiling}).
For details, see \ref{app:counting}.



\subsection{Spatial relations}
The \textbf{relations} piece has a single instrument and focuses on the ability of models to distinguish between different spatial relations. 
Foils differ from the original caption only by the replacement of a spatial preposition. 
As with plurals, the data
was sourced from the COCO 2017 validation split. To create foils, we first identified all
preposition sequences in captions (e.g., `in', `out of'). Foils were created 
by masking the prepositions and using  SpanBERT \cite{joshi-etal-2020-spanbert} to generate candidates of between 1--3 words in length. 
We keep SpanBERT candidates, which are spans whose lengths vary from 1 to 3, if they differ from the original preposition sequence, but exist in 
the dataset.
There are 535 instances after manual validation out of 614 proposed instances (cf. \S \ref{sec:foiling}),
and we ensure that prepositions are similarly distributed among captions and foils.
Full details are provided in \ref{app:relations}.

\subsection{Actions}\label{subsec:actions}
The \textbf{actions} piece has two instruments: i) \textbf{action replacement} and ii) \textbf{actant swap}.
They 
test a V\&L model's capability to i) identify whether an \textit{action} mentioned in the text matches the action seen in the image (e.g., `a man \underline{\gr{shouts}} / \underline{\red{smiles}} at a woman'), and ii) correctly identify the \textit{participants} of an action and the \textit{roles} they play (e.g., is it the man who is shouting or is it the woman, given the picture in Table \ref{tab:phenomena}?).

The SWiG dataset \cite{DBLP:conf/eccv/PrattYWFK20} contains 504 action verbs, and we
generate captions and foils from SWiG annotations of semantic roles and their fillers. For the action replacement piece, we exchange action verbs with other verbs from SWiG that fit the linguistic context as suggested by BERT. For the actant swap, we swap role fillers in the role annotations, hence generating action descriptions with inverted roles. 
Action replacement and actant swap have 648 (779) and 949 (1042) instances after (before) manual validation, respectively (cf. \S \ref{sec:foiling}).
See \ref{app:actions} for full details.

\subsection{Coreference}\label{subsec:coreference}
The \textbf{coreference} piece aims to uncover whether V\&L models are able to perform pronominal coreference resolution. It encompasses
cases where
i) the pronoun has a noun (phrase) antecedent
and pronoun and (noun) phrase are both grounded in the visual modality (`\underline{A woman} is driving a motorcycle. Is \underline{she} wearing a helmet?'), and cases where
ii) the pronoun refers to a region in the image or even to the  entire image (`Is \underline{this} outside?').

We create foils based on VisDial v1.0 \cite{visdial} with images from MSCOCO \cite{Lin-etal:2014:mscoco}.
VisDial captions and dialogues are
Q\&A sequences. We select image descriptions of the form [\emph{Caption. Question? Yes/No.}] where the question contains at least one pronoun. 
When foiling, we exchange the answer from \emph{yes} to \emph{no} and vice-versa (see Table \ref{tab:phenomena}). We ensure a 50-50\% balance between yes / no answers.

The coreference piece consists of two instruments: \textbf{coreference standard} originating from the VisDial train set and a small \textbf{coreference clean} set from the validation set, containing 708 (916) and 104 (141) examples after (before) manual validation, respectively (cf. \S \ref{sec:foiling}).\footnote{VisDial annotations are not available for the test set.} See  \ref{app:coref} for full details.

\section{\emph{Reliable} construction of \emph{valid} foils}
\label{sec:foiling}

In 
\dataset{}, an instance consisting of an image-caption-foil triple is considered {\em valid} if:
the foil minimally differs from the original caption;
the foil does not accurately describe the image;
and independent judges agree that the caption, but not the foil, is an accurate description of the image. 
We consider 
a \textit{foiling method} to be
more {\em reliable} the more it ensures that a generated foil does not substantially differ from a human caption regarding distributional and plausibility bias, and cannot be easily solved unimodally.  

In this section, we discuss automatic and manual means to reliably construct valid foils.
In this context, two types of bias are especially worthy of note:
distributional bias (\S\ref{sec:dist-bias}) and plausibility bias (\S\ref{subsec:why-unimodal}).
In \S\ref{subsec:nli} we discuss how we apply a natural language inference model to filter examples in our data pipeline, and \S\ref{sec:mturk} show how we manually validate \textit{all examples} in our benchmark.
Random samples from the final version of each instrument are shown in Tab.\ \ref{tab:existence-examples}--\ref{tab:coreference-examples}.

\subsection{Mitigating distributional bias}\label{sec:dist-bias}
A first form of bias is related to distributional imbalance between captions and foils (e.g., certain words or phrases having a high probability only in foils).
Previous foiling datasets
exhibit such
imbalance, enabling models to solve the task disregarding the image  \cite{madhyastha-etal-2019-vifidel}.
To mitigate this problem, for each phenomenon and throughout our data creation process, we ensure that the token \textit{frequency distributions} in correct and foiled captions are approximately the same
(cf. App. \ref{app:benchmark} and \ref{app:mturk}).



\subsection{Countering plausibility bias}\label{subsec:why-unimodal}


A second form of bias may arise from automatic procedures yielding foils that are implausible or unnatural, which can facilitate their detection. 
Often, \dataset{} pieces can be safely foiled by simple rules
(e.g., switching from existence to non-existence, or from singular to plural or vice versa).
However, with \textit{spatial relations} and \textit{actions}, a foil could be deemed unlikely given only the textual modality and independently of the image, e.g., `a man stands \underline{\red{under}} / \underline{\gr{on}} a chair'.
Such \textbf{plausibility biases} may be detected by large language models that incorporate commonsense knowledge \cite{petroni-etal-2019-language, wang2020language}, and we expect future V\&L models to exhibit similar capabilities.

To ensure that foiled and correct captions are similarly plausible,
we use language models such as BERT \cite{devlin2019bert} and SpanBERT \cite{joshi-etal-2020-spanbert} to suggest
replacements in our foiling functions.
Additionally, in the case of spatial relations and plurals, we also apply a grammaticality filter using GRUEN \cite{zhu-bhat-2020-gruen}. GRUEN was originally proposed to automatically score generated sentences based on discourse-level and grammatical properties. We use only the grammaticality component of GRUEN, and retain only foil candidates with a grammaticality score $\geq 0.8$.

Furthermore, we evaluate unimodal, language-only models on \dataset{} to 
verify
whether our benchmark could be solved by a multimodal model with strong linguistic capacities in \textbf{unimodal collapse}, whereby a model silently relies on a single modality within which biases are easier to exploit \cite{goyal2017making, shekhar-etal-2019-evaluating}.
By evaluating \dataset{} with unimodal models, we establish a baseline that V\&L models should exceed if we are to expect true multimodal integration.


\subsection{Filtering foils with NL Inference} \label{subsec:nli}
When constructing foils, we need to ensure that they \textit{fail} to describe the image. To test this automatically, we apply natural language inference (NLI) with the following rationale: We consider an image and its caption as a premise and its entailed hypothesis, respectively \citep[a similar rationale is applied in the visual entailment task;][]{Xie2019}. In addition, we consider the \textit{caption as premise} and the \textit{foil as its hypothesis}. 
If a NLI model predicts the foil to be entailed (E) by the caption, it cannot be a good foil since by transitivity it will give a truthful description of the image. 
By contrast, if the foil is predicted to contradict (C) or to be neutral (N) with respect to the caption, we take this as an indicator of a valid
(C) or a plausible (N) foil.\footnote{See the following examples from action replacement:\\
P: \textit{A mother scolds her son.} \\
H1: \textit{A mother encourages her son.} (C; good foil);\\
H2: \textit{A mother camps with her son.} (N; needs image control);\\
H3: \textit{A mother talks to her son.} (E; not a suitable foil)

If the NLI prediction is N,  we still need to check the image, since the description might happen to fit the image content.} 

We use the NLI model ALBERT \cite{lan2019albert} finetuned on the task (see Appendix \ref{app:filtering} for details).
Filtering with NLI was initially applied to \textit{relations, plurals} and \textit{actions}, on the grounds that foils in these pieces 
may induce
substantive changes to lexical content.\footnote{By contrast, existence and counting foils involve a more straightforward swap (e.g., between numerical quantities); similarly, coreference foils simply involve the replacement of a positive with a negative answer.} Following automatic labelling of caption-foil pairs, we manually validated a sample 
labelled as E, C or N. For \textit{relations} ($N=30$), labels were found to be near 100\% accurate with only 2 (0.06\%) errors overall. For \textit{plurals} ($N=60$, 50\% {\tt sg2pl} and 50\% {\tt pl2sg}), the error rate was also low, with 0 errors for C, 33\% errors for E and 11\% errors for N. Here,
a number of entailment errors were due to odd formulations arising from the automatic foiling process, whereas no such oddities were observed for C. We therefore include only foils labelled C in the final relations and plurals pieces. For \textit{actions}, the model labelled contradictions very accurately (0\% error) but was erroneous up to 97.1\% for E, meaning that a large number of valid foils would be spuriously excluded. To avoid reducing the dataset too much, we did not use NLI filtering for actions, but relied on human annotation as a final validity check.




\subsection{Manual evaluation of generated foils}\label{sec:mturk}
As a final step, the data for each instrument was submitted to a manual validation.
For each instance, annotators were shown the image, the caption and the foil. Caption and foil were numbered and displayed above each other to make differences more apparent, 
with differing elements highlighted in boldface (Fig.\ \ref{fig:mturk-example}, App. \ref{app:mturk}).
Annotators were not informed which text was the caption and which was the foil, and captions appeared first (numbered {\em 1}) 50\% of the time. The task was to determine which of the two texts accurately described what could be seen in the image. In each case, annotators had a forced choice between five options: a) the first, but not the second; b) the second, but not the first; c) both of them; d) neither of the two; and e) I cannot tell.

Each item was annotated by three individuals. The validation was conducted on Amazon Mechanical Turk with a fixed set of annotators who had qualified for the task.
For details see  App.\ \ref{app:mturk}.
For the final version of \dataset{}, we include instances which passed the following validation test: 
at least two out of three annotators identified the caption, but not the foil, as the text which accurately describes the image. Across all instruments, 87.7\% of the instances satisfied this criterion (min 77.3\%; max 94.6\%),
with 73.6\% of instances overall having a unanimous (3/3) decision that the caption, but not the foil, was an accurate description. We consider these figures high, suggesting that the automatic construction and filtering procedures yield foils which are likely to be valid, in the sense discussed in \S\ref{sec:foiling} above. 

We compute inter-annotator agreement for each instrument (Tab.~\ref{tab:mturk-results}). On the valid subset, agreement is low to medium (Krippendorff's $\alpha$: min=0.23, max=0.64, mean=0.42, sd=0.12). We note that there is considerable variation in the number of annotations made by individuals, and $\alpha$ is computed over 5 categories. Hence, this result cannot be straightforwardly interpreted as a ceiling of human performance for \dataset{}. However, $\alpha$ is higher for pieces on which models also perform better (e.g. existence, Foil-It!; cf. \S \ref{sec:benchmark}).


\section{Benchmarking with \dataset{}}\label{sec:benchmark}

\begin{table*}[t!]
    \small
    \centering
    \resizebox{\linewidth}{!}{
    \begin{tabular}{ l r r@{\hskip 0.2in}r@{\hskip 0.2in}rrr@{\hskip 0.2in}r@{\hskip 0.2in}rr@{\hskip 0.2in}rr@{\hskip 0.2in}r@{\hskip 0.2in}r }
        \toprule
        \multirow{2}{*}{\bf Metric} & \multirow{2}{*}{\bf Model} &
        \multicolumn{1}{c|}{\bf Existence} & \multicolumn{1}{c|}{\bf Plurality} & \multicolumn{3}{c|}{\bf Counting} & \multicolumn{1}{c|}{\bf Sp.rel.$\ddagger$} & \multicolumn{2}{c|}{\bf Action} & \multicolumn{2}{c|}{\bf Coreference} & \multicolumn{1}{c|}{\multirow{2}{*}{\bf Foil-it!}} & \multicolumn{1}{c}{\multirow{2}{*}{\bf Avg.}} \\
        && \multicolumn{1}{c|}{quantifiers} & \multicolumn{1}{c|}{number} & \multicolumn{1}{c}{balanced} & \multicolumn{1}{c}{sns.$\dagger$} & \multicolumn{1}{c|}{adv.$\dagger$} & \multicolumn{1}{c|}{relations} & \multicolumn{1}{c}{repl.$\dagger$} & \multicolumn{1}{c|}{actant swap} & \multicolumn{1}{c}{standard} & \multicolumn{1}{c|}{clean} &
        \multicolumn{1}{c|}{}\\ 
        \midrule
        & \multicolumn{1}{r}{Random} & 50.0 & 50.0 & 50.0 & 50.0 & 50.0 & 50.0 & 50.0 & 50.0 & 50.0 & 50.0 & 50.0 & 50.0 \\
        \midrule
        \multirow{7}{*}{$acc_r$}
        & \multirow{1}{*}{GPT1$^*$} & 61.8 & 53.1 & 51.2 & \redtable{48.7} & 69.5 & {\bf 77.2} & 65.4 & 72.2 & \redtable{45.6} & \redtable{45.2} & 77.5 & 60.7 \\
        & \multirow{1}{*}{GPT2$^*$} & 58.0 & 51.9 & 51.6 & \redtable{49.8} & \redtable{45.3} & 75.0 & 66.8 & {\bf 76.9} & 54.5 & \redtable{50.0} & 80.7 & 60.1 \\
        & \multirow{1}{*}{CLIP} & 66.9 & 56.2 & 62.1 & 62.5 & 57.5 & 64.3 & {\bf 75.6} & 68.6 & 52.1 & \redtable{49.7} & {\bf 88.8} & 64.0 \\
        & \multirow{1}{*}{LXMERT} & 78.6 & 64.4 & 62.2 & 69.2 & \redtable{42.6} & 60.2 & 54.8 & \redtable{45.8} & \redtable{46.8} & \redtable{44.2} & 87.1 & 59.6 \\
        & \multirow{1}{*}{ViLBERT} & 65.5 & 61.2 & 58.6 & 62.9 & 73.7 & 57.2 & 70.7 & 68.3 & \redtable{47.2} & \redtable{48.1} & 86.9 & 63.7 \\
        & \multirow{1}{*}{12-in-1} & {\bf 95.6} & {\bf 72.4} & {\bf 76.7} & {\bf 80.2} & {\bf 77.3} & 67.7 & 65.9 & 58.9 & {\bf 75.7} & {\bf 69.2} & 86.9 & {\bf 75.1} \\
        & \multirow{1}{*}{VisualBERT} & \redtable{39.7} & \redtable{45.7} & \redtable{48.2} & \redtable{48.2} & \redtable{50.0} & \redtable{39.7} & \redtable{49.2} & \redtable{44.4} & \redtable{49.5} & \redtable{47.6} & \redtable{48.5} & \redtable{46.4}\\
        \cmidrule{2-14}
        \multirow{4}{*}{$acc$}
        & \multirow{1}{*}{LXMERT} & 55.8 & 55.1 & 52.0 & 55.4 & \redtable{49.9} & 50.8 & 51.1 & \redtable{48.5} & \redtable{49.8} & \redtable{49.0} & 70.8 & 53.5 \\
        & \multirow{1}{*}{ViLBERT} & \redtable{2.4} & 50.3 & 50.7 & 50.6 & 51.8 & \redtable{49.9} & 52.6 & 50.4 & \redtable{50.0} & \redtable{50.0} & 55.9 & 51.3 \\
        & \multirow{1}{*}{12-in-1} & {\bf 89.0} & {\bf 62.0} & {\bf 64.9} & {\bf 69.2} & {\bf 66.7} & {\bf 53.4} & {\bf 57.3} & {\bf 52.2} & {\bf 54.4} & {\bf 54.3} & {\bf 71.5} & {\bf 63.2} \\
        & \multirow{1}{*}{VisualBERT} & \redtable{49.3} & \redtable{46.5} & \redtable{48.3} & \redtable{47.8} & \redtable{50.0} & \redtable{49.3} & \redtable{48.8} & \redtable{49.7} & \redtable{50.0} & \redtable{50.0} & \redtable{46.6} & \redtable{48.8} \\
        \cmidrule{2-14}
        \cmidrule{2-14}
        \multirow{4}{*}{min($p_c$, $p_f$)}
        & \multirow{1}{*}{LXMERT} & \redtable{41.6}  & \redtable{\bf 42.2} & 50.9 & \redtable{50.0} & \redtable{37.3} & \redtable{\bf 28.4} & \redtable{35.8} & \redtable{36.8} & \redtable{\bf 18.4} & \redtable{\bf 17.3} & {\bf 69.3} & \redtable{38.9} \\
        & \multirow{1}{*}{ViLBERT} & \redtable{47.9} & \redtable{2.1} & \redtable{24.4} & \redtable{24.7} & \redtable{17.5} & \redtable{1.5} & \redtable{11.9} & \redtable{7.1} & \redtable{1.3} & \redtable{1.9} & \redtable{12.9} & \redtable{13.9} \\
        & \multirow{1}{*}{12-in-1} & {\bf 85.0} & \redtable{33.4} & {\bf 64.3} & {\bf 61.7} & {\bf 59.5} & \redtable{13.3} & \redtable{\bf 47.8} & \redtable{\bf 37.6} & \redtable{15.8} & \redtable{13.5} & \redtable{48.8} & \redtable{\bf 43.7} \\
        & \multirow{1}{*}{VisualBERT} & \redtable{1.3}  & \redtable{0.3} & \redtable{0.0} & \redtable{0.0} & \redtable{0.0} & \redtable{1.3} & \redtable{0.0} & \redtable{0.0} & \redtable{0.0} & \redtable{0.0} & \redtable{0.2} & \redtable{0.3} \\
        \cmidrule{2-14}
        \multirow{4}{1cm}{$AUROC$ $\times 100$}
        & \multirow{1}{*}{LXMERT} & 60.5 & 57.3 & 53.8 & 57.7 & 50.5 & 51.9 & 52.1 & \redtable{47.6} & \redtable{49.8} & \redtable{49.5} & 76.9 & 55.2 \\
        & \multirow{1}{*}{ViLBERT} & 52.5 & 54.1 & 50.8 & 51.6 & 53.5 & 51.2 & 57.2 & {\bf 57.8} & \redtable{49.9} & \redtable{49.9} & 75.2 &  54.9 \\
        & \multirow{1}{*}{12-in-1} & {\bf 96.3} & {\bf 67.4} & {\bf 72.0} & {\bf 77.8} & {\bf 75.1} & {\bf 55.8} & {\bf 61.3} & 55.0 & {\bf 59.8} & {\bf 59.6} & {\bf 81.0} & {\bf 69.2} \\
        & \multirow{1}{*}{VisualBERT} & \redtable{28.9} & \redtable{29.0} & \redtable{24.5} & \redtable{16.5} & \redtable{20.9} & \redtable{45.2} & \redtable{17.7} & \redtable{36.3} & \redtable{45.3} & \redtable{46.3} & \redtable{28.5} & \redtable{30.8} \\
        \bottomrule
    \end{tabular}
    }
    \caption{Performance of unimodal and multimodal models on the \dataset{} benchmark according to different metrics. We bold-face the best overall result per metric, and underscore all results below (or at) the random baseline.
    $acc_r$ is a pairwise ranking accuracy where a prediction is considered correct if $p(caption,img) > p(foil,img)$.
    Precision $p_c$ and foil precision $p_f$ are \emph{competing} metrics where na\"{i}vely increasing one can decrease the other: therefore \emph{looking at the smaller number among the two gives a good intuition of how informed is a model prediction}.
    $\dagger${\bf sns.} Counting small numbers. {\bf adv.} Counting adversarial. {\bf repl.} Action replacement. $\ddagger$ {\bf Sp.rel.} Spatial relations. $^*$Unimodal text-only models that do not use images as input. CLIP is only tested in pairwise ranking mode (fn.\ 6).
    }
    \label{tab:bench_results}
\end{table*}

We propose \dataset{} as a task-independent, \emph{zero-shot} benchmark to assess the extent to which models learn to ground specific linguistic phenomena as a consequence of their pretraining (or fine-tuning). \dataset{} is built in the spirit of approaches such as Checklist \cite{ribeiro-etal-2020-beyond},
including
pairs consisting of captions and minimally edited foils.

The only requirement to
evaluate a model on our benchmark is:
\textit{i)} to have a binary classification head to predict whether an image-sentence pair is foiled,
or
\textit{ii)} to predict an image-sentence matching score between the image and the caption vs.\ the foil, returning the pair with the highest score. 
Systems reporting results on \dataset{} are expected to report any data used in model training prior to testing on \dataset{}, for comparability. 

\subsection{Benchmark Metrics} \label{subsec:metrics}
We employ five metrics\footnote{All metrics are defined in Appendix \ref{app:metrics}.} for evaluation:
overall {\bf accuracy} ($acc$) on all classes (foil and correct);
{\bf precision} ($p_c$) measuring how well models identify the \emph{correct} examples; {\bf foil precision} ($p_f$) measuring how well \emph{foiled} cases are identified;
{\bf pairwise ranking accuracy} ($acc_r$), which measures whether the image-sentence alignment score is greater for a correct image-text pair than for its foiled pair;
and
{\bf area under the receiver operating characteristic curve} (AUROC), which measures how well models distinguish correct vs. foiled examples across different prediction thresholds.
$acc_r$ is more permissive than $acc$ as 
it accepts model predictions if the score for a foil is lower than the caption's score.
Our main metrics are AUROC and $acc_r$. $acc_r$ gives results for a pair \textit{$\langle$image, caption$\rangle$} and \textit{$\langle$image, foil$\rangle$}. Both AUROC and $acc_r$ are well suited to evaluate minimally-edited pairs as neither uses a classification threshold. 
As for $p_c$ and $p_f$, since these are competing metrics where naively increasing one can decrease the other, we report the smaller of the two as an indicator of how informed model predictions are.
Since all instruments are implemented as a balanced binary classification, the random baseline is always 50\%.

\subsection{V\&L models}\label{subsec:vl_models}

We benchmark five V\&L models on VALSE: CLIP \citep{radford2021learning}, LXMERT \citep{tan-bansal-2019-lxmert}, ViLBERT \citep{lu2019vilbert}, ViLBERT 12-in-1 \citep{lu2020vilbert12in1}, and VisualBERT \citep{li2019visualbert}. These models have different architectures and are pretrained on a variety of tasks
with
different training data. We also benchmark two unimodal text-only models, GPT1 \citep{radford2018improving} and GPT2 \citep{radford2019language}. 
See Appendix \ref{app:models} for details on all these models used in our evaluation.

\paragraph{Unimodal models} GPT1 and GPT2
are
autoregressive language models pretrained on English text. We test whether \dataset{} is solvable by these unimodal models by computing the perplexity of the correct and foiled caption and \textit{predicting the entry with the lowest perplexity}. 
If the perplexity is higher for the foil, 
we take this as an indication that the foiled caption may suffer from \textbf{plausibility bias} or other linguistic biases (cf.\ \S \ref{subsec:why-unimodal}).

\subsection{Experiments and Results}
\label{subsec:experiments}
We test V\&L and unimodal models on \dataset{} in a zero-shot setting, 
and also evaluate on a number of correct captions and foils from the \textit{FOIL it!} dataset \cite{shekhar-etal-2017-foil} (cf. App. \ref{app:foilit} for details).
All results are listed in Table \ref{tab:bench_results}.


\paragraph{Unimodal results} For most instruments, unimodal results are close to random and hence do not signal strong linguistic or plausibility biases.
One exception is the original \textit{FOIL it!} dataset, in line with \citet{madhyastha-etal-2019-vifidel}'s findings.
Also the spatial relations (77.2\%), action replacement (66.8\%) and actant swap (76.9\%) instruments 
suggest plausibility biases in 
foils. Such biases are hard to avoid in automatic foil generation for actions due to the verb arguments' selectional restrictions, which are easily  violated when flipping role fillers, or replacing
the verb. 
Similar considerations hold for relations: though SpanBERT proposals are intended to aid selection of likely replacements for prepositions, plausibility issues arise with relatively rare argument-preposition combinations.

While these might be the first instruments in \dataset{} to be solved in the future, current V\&L models struggle to detect even blatant mismatches of actant swap, e.g., `A ball throws a tennis player.'
For \dataset{}, the unimodal scores will serve as a baseline for the pairwise accuracy of V\&L models.

\paragraph{Multimodal results} The best zero-shot results are achieved by ViLBERT 12-in-1 with the highest scores across the board, followed by ViLBERT, LXMERT, CLIP,\footnote{CLIP works in a contrastive fashion, therefore we report only $acc_r$ (cf. Appendix \ref{app:models} for details).} and finally VisualBERT. The latter obtains high $p_f$ but very low $p_c$ values---reflected in the $\min(p_c,p_f)$ scores---indicating that VisualBERT learned a heuristic that does not generalise
\cite[see][for similar observations with other models]{Hendricks2021}. We hypothesise that this is due to the way image-sentence alignment is framed in VisualBERT's pretraining: the model expects an image and a correct sentence $c_{1}$, and predicts whether a second sentence $c_2$ is a match.\footnote{$c_1$ is one of the 5 captions describing the relevant image in MSCOCO. During VisualBERT's pretraining, $c_2$ can be an alternative caption out of these 5, or a randomly drawn caption which does not describe the image. The pretraining task is to determine if $c_2$ correctly describes the image or not.}
During pretraining $c_{1}$ and $c_{2}$ are likely to differ in 
many ways, whereas in our setting, they are nearly identical.
This may bias the model against predicting foils, which would raise the value $p_f$.

Instruments centered on individual objects like existence and the \textit{FOIL it!} dataset are almost solved by ViLBERT 12-in-1, highlighting that models are capable of identifying named objects and their presence in images. However, none of the remaining pieces can be reliably solved in our adversarial foiling settings: i) distinguishing references to single vs.\ multiple objects 
or counting them in an image (plurality and counting); ii) correctly classifying a named spatial relation between objects in an image (relations); iii) distinguishing actions and identifying their participants, even if supported by preference biases (actions); or, iv) tracing multiple references to the same object in an image through the use of pronouns (coreference).

\paragraph{Correct vs. foil precision}
$p_c$ and $p_f$ show that V\&L models struggle to solve the phenomena in \dataset{}.
When a model achieves high precision on  correct captions $p_c$ this is often at the expense of very low precision on foiled captions $p_f$ 
(cf.\ ViLBERT), or vice-versa 
(cf.\ VisualBERT).
This suggests that such models are insensitive to \dataset{}'s inputs:
models that almost always predict a match will inflate $p_f$ at the expense of $p_c$.
$\min(p_c, p_f)$ reveals that
VisualBERT and ViLBERT perform poorly and below 
random baseline, and LXMERT 
close to or below it.
ViLBERT 12-in-1 performs strongly on existence, well on counting, but struggles on plurality, spatial relations, coreference, and actions. These tendencies we see reflected in our main metrics, $acc_r$ and AUROC.


\section{Conclusions and Future Work}\label{sec:conclusion}
We present the \dataset{} benchmark to help the community improve V\&L models by hard-testing their visual grounding capabilities through the lens of linguistic constructs.
Our experiments show that V\&L models identify named objects and their presence in images well (as shown by the existence piece), but struggle to ground 
their interdependence and relationships in visual scenes when forced to respect 
linguistic indicators.
We encourage the community to use \dataset{} for measuring 
progress towards V\&L models capable of true language grounding.
Furthermore, \dataset{} could be used as an indirect assessment of datasets, as models could be evaluated before and after training or fine-tuning to see if a dataset helps models improve on any of the aspects tested by \dataset{}.

\dataset{} is designed as a living benchmark. As future work we plan to extend it to further linguistic phenomena, and to source data from diverse V\&L datasets to cover more linguistic variability and image distributions.

\section*{Acknowledgments}
IC has received funding from the European Union’s Horizon 2020 research and innovation programme under the Marie Sk\l{}odowska-Curie grant agreement No 838188. 
AG and MC are supported by the European Union's Horizon 2020 research and innovation Programme under the Marie Sk\l{}odowska-Curie grant agreement No 860621. 
This collaboration was facilitated by the Multi3Generation COST Action CA18231.

\bibliography{anthology,refs}

\begin{thebibliography}{67}
\expandafter\ifx\csname natexlab\endcsname\relax\def\natexlab#1{#1}\fi

\bibitem[{Agarwal et~al.(2020)Agarwal, Shetty, and Fritz}]{agarwal2020towards}
Vedika Agarwal, Rakshith Shetty, and Mario Fritz. 2020.
\newblock Towards causal vqa: Revealing and reducing spurious correlations by
  invariant and covariant semantic editing.
\newblock In \emph{Proceedings of the IEEE/CVF Conference on Computer Vision
  and Pattern Recognition}, pages 9690--9698.

\bibitem[{Akula et~al.(2020)Akula, Gella, Al-Onaizan, Zhu, and
  Reddy}]{akula-etal-2020-words}
Arjun Akula, Spandana Gella, Yaser Al-Onaizan, Song-Chun Zhu, and Siva Reddy.
  2020.
\newblock \href {https://doi.org/10.18653/v1/2020.acl-main.586} {Words aren{'}t
  enough, their order matters: On the robustness of grounding visual referring
  expressions}.
\newblock In \emph{Proceedings of the 58th Annual Meeting of the Association
  for Computational Linguistics}, pages 6555--6565, Online. Association for
  Computational Linguistics.

\bibitem[{Bernardi and Pezzelle(2021)}]{Bernardi2021}
Raffaella Bernardi and Sandro Pezzelle. 2021.
\newblock \href {https://doi.org/10.1111/lnc3.12417} {{Linguistic issues behind
  visual question answering}}.
\newblock \emph{Language and Linguistics Compass}, 15(6):1--25.

\bibitem[{Bitton et~al.(2021)Bitton, Stanovsky, Schwartz, and
  Elhadad}]{bitton-etal-2021-automatic}
Yonatan Bitton, Gabriel Stanovsky, Roy Schwartz, and Michael Elhadad. 2021.
\newblock \href {https://www.aclweb.org/anthology/2021.naacl-main.9} {Automatic
  generation of contrast sets from scene graphs: Probing the compositional
  consistency of {GQA}}.
\newblock In \emph{Proceedings of the 2021 Conference of the North American
  Chapter of the Association for Computational Linguistics: Human Language
  Technologies}, pages 94--105, Online. Association for Computational
  Linguistics.

\bibitem[{Bowman et~al.(2015)Bowman, Angeli, Potts, and
  Manning}]{bowman2015large}
Samuel~R. Bowman, Gabor Angeli, Christopher Potts, and Christopher~D. Manning.
  2015.
\newblock \href {https://doi.org/10.18653/v1/D15-1075} {A large annotated
  corpus for learning natural language inference}.
\newblock In \emph{Proceedings of the 2015 Conference on Empirical Methods in
  Natural Language Processing}, pages 632--642, Lisbon, Portugal. Association
  for Computational Linguistics.

\bibitem[{Cao et~al.(2020)Cao, Gan, Cheng, Yu, Chen, and Liu}]{cao2020behind}
Jize Cao, Zhe Gan, Yu~Cheng, Licheng Yu, Yen-Chun Chen, and Jingjing Liu. 2020.
\newblock Behind the scene: Revealing the secrets of pre-trained
  vision-and-language models.
\newblock \emph{arXiv preprint arXiv:2005.07310}.

\bibitem[{Chen et~al.(2015)Chen, Fang, Lin, Vedantam, Zitnick, Gupta, and
  Doll}]{Chen2015}
Xinlei Chen, Hao Fang, Tsung-yi Lin, Ramakrishna Vedantam, C~Lawrence Zitnick,
  Saurabh Gupta, and Piotr Doll. 2015.
\newblock \href {http://arxiv.org/abs/arXiv:1504.00325v2} {{Microsoft COCO
  Captions : Data Collection and Evaluation Server}}.
\newblock \emph{arXiv}, 1504.00325:1--7.

\bibitem[{Chen et~al.(2020)Chen, Li, Yu, Kholy, Ahmed, Gan, Cheng, and
  Liu}]{chen2020uniter}
Yen-Chun Chen, Linjie Li, Licheng Yu, Ahmed~El Kholy, Faisal Ahmed, Zhe Gan,
  Yu~Cheng, and Jingjing Liu. 2020.
\newblock Uniter: Universal image-text representation learning.
\newblock In \emph{ECCV}.

\bibitem[{Cirik et~al.(2018)Cirik, Morency, and
  Berg-Kirkpatrick}]{cirik-etal-2018-visual}
Volkan Cirik, Louis-Philippe Morency, and Taylor Berg-Kirkpatrick. 2018.
\newblock \href {https://doi.org/10.18653/v1/N18-2123} {Visual referring
  expression recognition: What do systems actually learn?}
\newblock In \emph{Proceedings of the 2018 Conference of the North {A}merican
  Chapter of the Association for Computational Linguistics: Human Language
  Technologies, Volume 2 (Short Papers)}, pages 781--787, New Orleans,
  Louisiana. Association for Computational Linguistics.

\bibitem[{Das et~al.(2017)Das, Kottur, Gupta, Singh, Yadav, Moura, Parikh, and
  Batra}]{visdial}
Abhishek Das, Satwik Kottur, Khushi Gupta, Avi Singh, Deshraj Yadav,
  Jos\'e~M.F. Moura, Devi Parikh, and Dhruv Batra. 2017.
\newblock {V}isual {D}ialog.
\newblock In \emph{Proceedings of the IEEE Conference on Computer Vision and
  Pattern Recognition (CVPR)}.

\bibitem[{Devlin et~al.(2019)Devlin, Chang, Lee, and
  Toutanova}]{devlin2019bert}
Jacob Devlin, Ming-Wei Chang, Kenton Lee, and Kristina Toutanova. 2019.
\newblock \href {https://doi.org/10.18653/v1/N19-1423} {{BERT}: Pre-training of
  deep bidirectional transformers for language understanding}.
\newblock In \emph{Proceedings of the 2019 Conference of the North {A}merican
  Chapter of the Association for Computational Linguistics: Human Language
  Technologies, Volume 1 (Long and Short Papers)}, pages 4171--4186,
  Minneapolis, Minnesota. Association for Computational Linguistics.

\bibitem[{Fellbaum(1998)}]{wordnet}
Christiane Fellbaum. 1998.
\newblock \emph{WordNet: An Electronic Lexical Database}.
\newblock Bradford Books.

\bibitem[{Gardner et~al.(2020)Gardner, Artzi, Basmov, Berant, Bogin, Chen,
  Dasigi, Dua, Elazar, Gottumukkala, Gupta, Hajishirzi, Ilharco, Khashabi, Lin,
  Liu, Liu, Mulcaire, Ning, Singh, Smith, Subramanian, Tsarfaty, Wallace,
  Zhang, and Zhou}]{gardner-etal-2020-evaluating}
Matt Gardner, Yoav Artzi, Victoria Basmov, Jonathan Berant, Ben Bogin, Sihao
  Chen, Pradeep Dasigi, Dheeru Dua, Yanai Elazar, Ananth Gottumukkala, Nitish
  Gupta, Hannaneh Hajishirzi, Gabriel Ilharco, Daniel Khashabi, Kevin Lin,
  Jiangming Liu, Nelson~F. Liu, Phoebe Mulcaire, Qiang Ning, Sameer Singh,
  Noah~A. Smith, Sanjay Subramanian, Reut Tsarfaty, Eric Wallace, Ally Zhang,
  and Ben Zhou. 2020.
\newblock \href {https://doi.org/10.18653/v1/2020.findings-emnlp.117}
  {Evaluating models{'} local decision boundaries via contrast sets}.
\newblock In \emph{Findings of the Association for Computational Linguistics:
  EMNLP 2020}, pages 1307--1323, Online. Association for Computational
  Linguistics.

\bibitem[{Garg et~al.(2019)Garg, Perot, Limtiaco, Taly, Chi, and
  Beutel}]{garg-etal-2019-counterfactual}
Sahaj Garg, Vincent Perot, Nicole Limtiaco, Ankur Taly, Ed~H. Chi, and Alex
  Beutel. 2019.
\newblock \href {https://doi.org/10.1145/3306618.3317950} {Counterfactual
  fairness in text classification through robustness}.
\newblock In \emph{Proceedings of the 2019 AAAI/ACM Conference on AI, Ethics,
  and Society}, AIES '19, page 219–226, New York, NY, USA. Association for
  Computing Machinery.

\bibitem[{Gatt and Reiter(2009)}]{gatt-reiter-2009-simplenlg}
Albert Gatt and Ehud Reiter. 2009.
\newblock \href {https://www.aclweb.org/anthology/W09-0613} {{S}imple{NLG}: A
  realisation engine for practical applications}.
\newblock In \emph{Proceedings of the 12th {E}uropean Workshop on Natural
  Language Generation ({ENLG} 2009)}, pages 90--93, Athens, Greece. Association
  for Computational Linguistics.

\bibitem[{Goh et~al.(2021)Goh, †, †, Carter, Petrov, Schubert, Radford, and
  Olah}]{goh2021multimodal}
Gabriel Goh, Nick~Cammarata †, Chelsea~Voss †, Shan Carter, Michael Petrov,
  Ludwig Schubert, Alec Radford, and Chris Olah. 2021.
\newblock \href {https://doi.org/10.23915/distill.00030} {Multimodal neurons in
  artificial neural networks}.
\newblock \emph{Distill}.
\newblock Https://distill.pub/2021/multimodal-neurons.

\bibitem[{Gokhale et~al.(2020)Gokhale, Banerjee, Baral, and
  Yang}]{gokhale-etal-2020-mutant}
Tejas Gokhale, Pratyay Banerjee, Chitta Baral, and Yezhou Yang. 2020.
\newblock \href {https://doi.org/10.18653/v1/2020.emnlp-main.63} {{MUTANT}: A
  training paradigm for out-of-distribution generalization in visual question
  answering}.
\newblock In \emph{Proceedings of the 2020 Conference on Empirical Methods in
  Natural Language Processing (EMNLP)}, pages 878--892, Online. Association for
  Computational Linguistics.

\bibitem[{Goyal et~al.(2017)Goyal, Khot, Summers-Stay, Batra, and
  Parikh}]{goyal2017making}
Yash Goyal, Tejas Khot, Douglas Summers-Stay, Dhruv Batra, and Devi Parikh.
  2017.
\newblock Making the v in vqa matter: Elevating the role of image understanding
  in visual question answering.
\newblock In \emph{Proceedings of the IEEE Conference on Computer Vision and
  Pattern Recognition}, pages 6904--6913.

\bibitem[{Gururangan et~al.(2018)Gururangan, Swayamdipta, Levy, Schwartz,
  Bowman, and Smith}]{gururangan-etal-2018-annotation}
Suchin Gururangan, Swabha Swayamdipta, Omer Levy, Roy Schwartz, Samuel Bowman,
  and Noah~A. Smith. 2018.
\newblock \href {https://doi.org/10.18653/v1/N18-2017} {Annotation artifacts in
  natural language inference data}.
\newblock In \emph{Proceedings of the 2018 Conference of the North {A}merican
  Chapter of the Association for Computational Linguistics: Human Language
  Technologies, Volume 2 (Short Papers)}, pages 107--112, New Orleans,
  Louisiana. Association for Computational Linguistics.

\bibitem[{He et~al.(2021)He, Wang, Miao, and Sun}]{HE2021104194}
Feijuan He, Yaxian Wang, Xianglin Miao, and Xia Sun. 2021.
\newblock \href {https://doi.org/https://doi.org/10.1016/j.imavis.2021.104194}
  {Interpretable visual reasoning: A survey}.
\newblock \emph{Image and Vision Computing}, 112:104194.

\bibitem[{Hendricks and Nematzadeh(2021)}]{Hendricks2021}
Lisa~Anne Hendricks and Aida Nematzadeh. 2021.
\newblock \href {http://arxiv.org/abs/2106.09141} {{Probing Image-Language
  Transformers for Verb Understanding}}.
\newblock \emph{arXiv}, 2106.09141.

\bibitem[{Jia and Liang(2017)}]{jia-liang-2017-adversarial}
Robin Jia and Percy Liang. 2017.
\newblock \href {https://doi.org/10.18653/v1/D17-1215} {Adversarial examples
  for evaluating reading comprehension systems}.
\newblock In \emph{Proceedings of the 2017 Conference on Empirical Methods in
  Natural Language Processing}, pages 2021--2031, Copenhagen, Denmark.
  Association for Computational Linguistics.

\bibitem[{Jia et~al.(2019)Jia, Raghunathan, G{\"o}ksel, and
  Liang}]{jia-etal-2019-certified}
Robin Jia, Aditi Raghunathan, Kerem G{\"o}ksel, and Percy Liang. 2019.
\newblock \href {https://doi.org/10.18653/v1/D19-1423} {Certified robustness to
  adversarial word substitutions}.
\newblock In \emph{Proceedings of the 2019 Conference on Empirical Methods in
  Natural Language Processing and the 9th International Joint Conference on
  Natural Language Processing (EMNLP-IJCNLP)}, pages 4129--4142, Hong Kong,
  China. Association for Computational Linguistics.

\bibitem[{Joshi et~al.(2020)Joshi, Chen, Liu, Weld, Zettlemoyer, and
  Levy}]{joshi-etal-2020-spanbert}
Mandar Joshi, Danqi Chen, Yinhan Liu, Daniel~S. Weld, Luke Zettlemoyer, and
  Omer Levy. 2020.
\newblock \href {https://doi.org/10.1162/tacl_a_00300} {{S}pan{BERT}: Improving
  pre-training by representing and predicting spans}.
\newblock \emph{Transactions of the Association for Computational Linguistics},
  8:64--77.

\bibitem[{Kafle et~al.(2019)Kafle, Shrestha, and
  Kanan}]{10.3389/frai.2019.00028}
Kushal Kafle, Robik Shrestha, and Christopher Kanan. 2019.
\newblock \href {https://doi.org/10.3389/frai.2019.00028} {Challenges and
  prospects in vision and language research}.
\newblock \emph{Frontiers in Artificial Intelligence}, 2:28.

\bibitem[{Kim et~al.(2021)Kim, Son, and Kim}]{kim2021vilt}
Wonjae Kim, Bokyung Son, and Ildoo Kim. 2021.
\newblock Vilt: Vision-and-language transformer without convolution or region
  supervision.
\newblock \emph{arXiv preprint arXiv:2102.03334}.

\bibitem[{Lan et~al.(2020)Lan, Chen, Goodman, Gimpel, Sharma, and
  Soricut}]{lan2019albert}
Zhenzhong Lan, Mingda Chen, Sebastian Goodman, Kevin Gimpel, Piyush Sharma, and
  Radu Soricut. 2020.
\newblock \href {https://openreview.net/forum?id=H1eA7AEtvS} {Albert: A lite
  bert for self-supervised learning of language representations}.
\newblock In \emph{International Conference on Learning Representations}.

\bibitem[{Levesque et~al.(2012)Levesque, Davis, and
  Morgenstern}]{levesque2012winograd}
Hector Levesque, Ernest Davis, and Leora Morgenstern. 2012.
\newblock The winograd schema challenge.
\newblock In \emph{Thirteenth International Conference on the Principles of
  Knowledge Representation and Reasoning}.

\bibitem[{Li et~al.(2020{\natexlab{a}})Li, Duan, Fang, Gong, and
  Jiang}]{Li-etal-2020unicodervl}
Gen Li, Nan Duan, Yuejian Fang, Ming Gong, and Daxin Jiang. 2020{\natexlab{a}}.
\newblock \href {https://aaai.org/ojs/index.php/AAAI/article/view/6795}
  {Unicoder-vl: {A} universal encoder for vision and language by cross-modal
  pre-training}.
\newblock In \emph{The Thirty-Fourth {AAAI} Conference on Artificial
  Intelligence, {AAAI} 2020, The Thirty-Second Innovative Applications of
  Artificial Intelligence Conference, {IAAI} 2020, The Tenth {AAAI} Symposium
  on Educational Advances in Artificial Intelligence, {EAAI} 2020, New York,
  NY, USA, February 7-12, 2020}, pages 11336--11344. {AAAI} Press.

\bibitem[{Li et~al.(2021)Li, Lei, Gan, Yu, Chen, Pillai, Cheng, Zhou, Wang,
  Wang et~al.}]{li2021value}
Linjie Li, Jie Lei, Zhe Gan, Licheng Yu, Yen-Chun Chen, Rohit Pillai, Yu~Cheng,
  Luowei Zhou, Xin~Eric Wang, William~Yang Wang, et~al. 2021.
\newblock Value: A multi-task benchmark for video-and-language understanding
  evaluation.
\newblock \emph{arXiv preprint arXiv:2106.04632}.

\bibitem[{Li et~al.(2019)Li, Yatskar, Yin, Hsieh, and Chang}]{li2019visualbert}
Liunian~Harold Li, Mark Yatskar, Da~Yin, Cho-Jui Hsieh, and Kai-Wei Chang.
  2019.
\newblock Visualbert: A simple and performant baseline for vision and language.
\newblock In \emph{Arxiv}.

\bibitem[{Li et~al.(2020{\natexlab{b}})Li, Yin, Li, Zhang, Hu, Zhang, Wang, Hu,
  Dong, Wei et~al.}]{li2020oscar}
Xiujun Li, Xi~Yin, Chunyuan Li, Pengchuan Zhang, Xiaowei Hu, Lei Zhang, Lijuan
  Wang, Houdong Hu, Li~Dong, Furu Wei, et~al. 2020{\natexlab{b}}.
\newblock Oscar: Object-semantics aligned pre-training for vision-language
  tasks.
\newblock In \emph{European Conference on Computer Vision}, pages 121--137.
  Springer.

\bibitem[{Lin et~al.(2014)Lin, Maire, Belongie, Hays, Perona, Ramanan,
  Doll{\'a}r, and Zitnick}]{Lin-etal:2014:mscoco}
Tsung-Yi Lin, Michael Maire, Serge Belongie, James Hays, Pietro Perona, Deva
  Ramanan, Piotr Doll{\'a}r, and C.~Lawrence Zitnick. 2014.
\newblock Microsoft coco: Common objects in context.
\newblock In \emph{Computer Vision -- ECCV 2014}, pages 740--755, Cham.
  Springer International Publishing.

\bibitem[{Lourie et~al.(2021)Lourie, Le~Bras, Bhagavatula, and
  Choi}]{lourie2021unicorn}
Nicholas Lourie, Ronan Le~Bras, Chandra Bhagavatula, and Yejin Choi. 2021.
\newblock Unicorn on rainbow: A universal commonsense reasoning model on a new
  multitask benchmark.
\newblock In \emph{Proceedings of the AAAI Conference on Artificial
  Intelligence}, 15, pages 13480--13488.

\bibitem[{Lu et~al.(2019)Lu, Batra, Parikh, and Lee}]{lu2019vilbert}
Jiasen Lu, Dhruv Batra, Devi Parikh, and Stefan Lee. 2019.
\newblock Vilbert: Pretraining task-agnostic visiolinguistic representations
  for vision-and-language tasks.
\newblock In \emph{Advances in Neural Information Processing Systems}, pages
  13--23.

\bibitem[{Lu et~al.(2020)Lu, Goswami, Rohrbach, Parikh, and
  Lee}]{lu2020vilbert12in1}
Jiasen Lu, Vedanuj Goswami, Marcus Rohrbach, Devi Parikh, and Stefan Lee. 2020.
\newblock 12-in-1: Multi-task vision and language representation learning.
\newblock In \emph{The IEEE/CVF Conference on Computer Vision and Pattern
  Recognition (CVPR)}.

\bibitem[{Madhyastha et~al.(2019)Madhyastha, Wang, and
  Specia}]{madhyastha-etal-2019-vifidel}
Pranava Madhyastha, Josiah Wang, and Lucia Specia. 2019.
\newblock \href {https://doi.org/10.18653/v1/P19-1654} {{VIFIDEL}: Evaluating
  the visual fidelity of image descriptions}.
\newblock In \emph{Proceedings of the 57th Annual Meeting of the Association
  for Computational Linguistics}, pages 6539--6550, Florence, Italy.
  Association for Computational Linguistics.

\bibitem[{Madhyastha et~al.(2018)Madhyastha, Wang, and
  Specia}]{madhyastha-etal-2018-defoiling}
Pranava~Swaroop Madhyastha, Josiah Wang, and Lucia Specia. 2018.
\newblock \href {https://doi.org/10.18653/v1/N18-2069} {Defoiling foiled image
  captions}.
\newblock In \emph{Proceedings of the 2018 Conference of the North {A}merican
  Chapter of the Association for Computational Linguistics: Human Language
  Technologies, Volume 2 (Short Papers)}, pages 433--438, New Orleans,
  Louisiana. Association for Computational Linguistics.

\bibitem[{Nie et~al.(2019)Nie, Chen, and Bansal}]{nie2019combining}
Yixin Nie, Haonan Chen, and Mohit Bansal. 2019.
\newblock Combining fact extraction and verification with neural semantic
  matching networks.
\newblock In \emph{Association for the Advancement of Artificial Intelligence
  ({AAAI})}.

\bibitem[{Nie et~al.(2020)Nie, Williams, Dinan, Bansal, Weston, and
  Kiela}]{nie2019adversarial}
Yixin Nie, Adina Williams, Emily Dinan, Mohit Bansal, Jason Weston, and Douwe
  Kiela. 2020.
\newblock \href {https://doi.org/10.18653/v1/2020.acl-main.441} {Adversarial
  {NLI}: A new benchmark for natural language understanding}.
\newblock In \emph{Proceedings of the 58th Annual Meeting of the Association
  for Computational Linguistics}, pages 4885--4901, Online. Association for
  Computational Linguistics.

\bibitem[{Parcalabescu et~al.(2021)Parcalabescu, Gatt, Frank, and
  Calixto}]{parcalabescu2021seeing}
Letitia Parcalabescu, Albert Gatt, Anette Frank, and Iacer Calixto. 2021.
\newblock \href {https://www.aclweb.org/anthology/2021.mmsr-1.4} {Seeing past
  words: Testing the cross-modal capabilities of pretrained v\&l models on
  counting tasks}.
\newblock In \emph{Proceedings of the 1st Workshop on Multimodal Semantic
  Representations (MMSR)}, pages 32--44, Groningen, Netherlands (Online).
  Association for Computational Linguistics.

\bibitem[{Petroni et~al.(2019)Petroni, Rockt{\"a}schel, Riedel, Lewis, Bakhtin,
  Wu, and Miller}]{petroni-etal-2019-language}
Fabio Petroni, Tim Rockt{\"a}schel, Sebastian Riedel, Patrick Lewis, Anton
  Bakhtin, Yuxiang Wu, and Alexander Miller. 2019.
\newblock \href {https://doi.org/10.18653/v1/D19-1250} {Language models as
  knowledge bases?}
\newblock In \emph{Proceedings of the 2019 Conference on Empirical Methods in
  Natural Language Processing and the 9th International Joint Conference on
  Natural Language Processing (EMNLP-IJCNLP)}, pages 2463--2473, Hong Kong,
  China. Association for Computational Linguistics.

\bibitem[{Plummer et~al.(2015)Plummer, Wang, Cervantes, Caicedo, Hockenmaier,
  and Lazebnik}]{plummer2015flickr30k}
Bryan~A Plummer, Liwei Wang, Chris~M Cervantes, Juan~C Caicedo, Julia
  Hockenmaier, and Svetlana Lazebnik. 2015.
\newblock Flickr30k entities: Collecting region-to-phrase correspondences for
  richer image-to-sentence models.
\newblock In \emph{Proceedings of the IEEE international conference on computer
  vision}, pages 2641--2649.

\bibitem[{Prabhakaran et~al.(2021)Prabhakaran, Davani, and
  Díaz}]{prabhakaran2021releasing}
Vinodkumar Prabhakaran, Aida~Mostafazadeh Davani, and Mark Díaz. 2021.
\newblock \href {http://arxiv.org/abs/2110.05699} {On releasing annotator-level
  labels and information in datasets}.

\bibitem[{Pratt et~al.(2020)Pratt, Yatskar, Weihs, Farhadi, and
  Kembhavi}]{DBLP:conf/eccv/PrattYWFK20}
Sarah Pratt, Mark Yatskar, Luca Weihs, Ali Farhadi, and Aniruddha Kembhavi.
  2020.
\newblock \href {https://doi.org/10.1007/978-3-030-58548-8\_19} {Grounded
  situation recognition}.
\newblock In \emph{Computer Vision - {ECCV} 2020 - 16th European Conference},
  pages 314--332.

\bibitem[{Radford et~al.(2021)Radford, Kim, Hallacy, Ramesh, Goh, Agarwal,
  Sastry, Askell, Mishkin, Clark et~al.}]{radford2021learning}
Alec Radford, Jong~Wook Kim, Chris Hallacy, Aditya Ramesh, Gabriel Goh,
  Sandhini Agarwal, Girish Sastry, Amanda Askell, Pamela Mishkin, Jack Clark,
  et~al. 2021.
\newblock Learning transferable visual models from natural language
  supervision.
\newblock \emph{arXiv preprint arXiv:2103.00020}.

\bibitem[{Radford et~al.(2018)Radford, Narasimhan, Salimans, and
  Sutskever}]{radford2018improving}
Alec Radford, Karthik Narasimhan, Tim Salimans, and Ilya Sutskever. 2018.
\newblock Improving language understanding by generative pre-training.

\bibitem[{Radford et~al.(2019)Radford, Wu, Child, Luan, Amodei, and
  Sutskever}]{radford2019language}
Alec Radford, Jeffrey Wu, Rewon Child, David Luan, Dario Amodei, and Ilya
  Sutskever. 2019.
\newblock Language models are unsupervised multitask learners.
\newblock \emph{OpenAI blog}, 1(8):9.

\bibitem[{Ribeiro et~al.(2020)Ribeiro, Wu, Guestrin, and
  Singh}]{ribeiro-etal-2020-beyond}
Marco~Tulio Ribeiro, Tongshuang Wu, Carlos Guestrin, and Sameer Singh. 2020.
\newblock \href {https://doi.org/10.18653/v1/2020.acl-main.442} {Beyond
  accuracy: Behavioral testing of {NLP} models with {C}heck{L}ist}.
\newblock In \emph{Proceedings of the 58th Annual Meeting of the Association
  for Computational Linguistics}, pages 4902--4912, Online. Association for
  Computational Linguistics.

\bibitem[{Rosenberg et~al.(2021)Rosenberg, Gat, Feder, and
  Reichart}]{rosenberg-etal-2021-vqa}
Daniel Rosenberg, Itai Gat, Amir Feder, and Roi Reichart. 2021.
\newblock \href {https://doi.org/10.18653/v1/2021.acl-short.10} {Are {VQA}
  systems {RAD}? {M}easuring robustness to augmented data with focused
  interventions}.
\newblock In \emph{Proceedings of the 59th Annual Meeting of the Association
  for Computational Linguistics and the 11th International Joint Conference on
  Natural Language Processing (Volume 2: Short Papers)}, pages 61--70, Online.
  Association for Computational Linguistics.

\bibitem[{Sharma et~al.(2018)Sharma, Ding, Goodman, and
  Soricut}]{sharma2018conceptual}
Piyush Sharma, Nan Ding, Sebastian Goodman, and Radu Soricut. 2018.
\newblock \href {https://doi.org/10.18653/v1/P18-1238} {Conceptual captions: A
  cleaned, hypernymed, image alt-text dataset for automatic image captioning}.
\newblock In \emph{Proceedings of the 56th Annual Meeting of the Association
  for Computational Linguistics (Volume 1: Long Papers)}, pages 2556--2565,
  Melbourne, Australia. Association for Computational Linguistics.

\bibitem[{Shekhar et~al.(2017{\natexlab{a}})Shekhar, Pezzelle, Herbelot, Nabi,
  Sangineto, and Bernardi}]{shekhar-etal-2017-vision}
Ravi Shekhar, Sandro Pezzelle, Aur{\'e}lie Herbelot, Moin Nabi, Enver
  Sangineto, and Raffaella Bernardi. 2017{\natexlab{a}}.
\newblock \href {https://www.aclweb.org/anthology/W17-6938} {Vision and
  language integration: Moving beyond objects}.
\newblock In \emph{{IWCS} 2017 {---} 12th International Conference on
  Computational Semantics {---} Short papers}.

\bibitem[{Shekhar et~al.(2017{\natexlab{b}})Shekhar, Pezzelle, Klimovich,
  Herbelot, Nabi, Sangineto, and Bernardi}]{shekhar-etal-2017-foil}
Ravi Shekhar, Sandro Pezzelle, Yauhen Klimovich, Aur{\'e}lie Herbelot, Moin
  Nabi, Enver Sangineto, and Raffaella Bernardi. 2017{\natexlab{b}}.
\newblock \href {https://doi.org/10.18653/v1/P17-1024} {{FOIL} it! find one
  mismatch between image and language caption}.
\newblock In \emph{Proceedings of the 55th Annual Meeting of the Association
  for Computational Linguistics (Volume 1: Long Papers)}, pages 255--265,
  Vancouver, Canada. Association for Computational Linguistics.

\bibitem[{Shekhar et~al.(2019{\natexlab{a}})Shekhar, Takmaz, Fern{\'a}ndez, and
  Bernardi}]{shekhar-etal-2019-evaluating}
Ravi Shekhar, Ece Takmaz, Raquel Fern{\'a}ndez, and Raffaella Bernardi.
  2019{\natexlab{a}}.
\newblock \href {https://doi.org/10.18653/v1/W19-0418} {Evaluating the
  representational hub of language and vision models}.
\newblock In \emph{Proceedings of the 13th International Conference on
  Computational Semantics - Long Papers}, pages 211--222, Gothenburg, Sweden.
  Association for Computational Linguistics.

\bibitem[{Shekhar et~al.(2019{\natexlab{b}})Shekhar, Venkatesh,
  Baumg{\"a}rtner, Bruni, Plank, Bernardi, and
  Fern{\'a}ndez}]{shekhar-etal-2019-beyond}
Ravi Shekhar, Aashish Venkatesh, Tim Baumg{\"a}rtner, Elia Bruni, Barbara
  Plank, Raffaella Bernardi, and Raquel Fern{\'a}ndez. 2019{\natexlab{b}}.
\newblock \href {https://doi.org/10.18653/v1/N19-1265} {Beyond task success: A
  closer look at jointly learning to see, ask, and {G}uess{W}hat}.
\newblock In \emph{Proceedings of the 2019 Conference of the North {A}merican
  Chapter of the Association for Computational Linguistics: Human Language
  Technologies, Volume 1 (Long and Short Papers)}, pages 2578--2587,
  Minneapolis, Minnesota. Association for Computational Linguistics.

\bibitem[{Su et~al.(2020)Su, Zhu, Cao, Li, Lu, Wei, and Dai}]{Su2020VL-BERT}
Weijie Su, Xizhou Zhu, Yue Cao, Bin Li, Lewei Lu, Furu Wei, and Jifeng Dai.
  2020.
\newblock \href {https://openreview.net/forum?id=SygXPaEYvH} {Vl-bert:
  Pre-training of generic visual-linguistic representations}.
\newblock In \emph{International Conference on Learning Representations}.

\bibitem[{Suhr et~al.(2019)Suhr, Zhou, Zhang, Zhang, Bai, and
  Artzi}]{suhr2018corpus}
Alane Suhr, Stephanie Zhou, Ally Zhang, Iris Zhang, Huajun Bai, and Yoav Artzi.
  2019.
\newblock \href {https://doi.org/10.18653/v1/P19-1644} {A corpus for reasoning
  about natural language grounded in photographs}.
\newblock In \emph{Proceedings of the 57th Annual Meeting of the Association
  for Computational Linguistics}, pages 6418--6428, Florence, Italy.
  Association for Computational Linguistics.

\bibitem[{Tan and Bansal(2019)}]{tan-bansal-2019-lxmert}
Hao Tan and Mohit Bansal. 2019.
\newblock \href {https://doi.org/10.18653/v1/D19-1514} {{LXMERT}: Learning
  cross-modality encoder representations from transformers}.
\newblock In \emph{Proceedings of the 2019 Conference on Empirical Methods in
  Natural Language Processing and the 9th International Joint Conference on
  Natural Language Processing (EMNLP-IJCNLP)}, pages 5100--5111, Hong Kong,
  China. Association for Computational Linguistics.

\bibitem[{Wallace et~al.(2020)Wallace, Gardner, and
  Singh}]{wallace-etal-2020-interpreting}
Eric Wallace, Matt Gardner, and Sameer Singh. 2020.
\newblock \href {https://doi.org/10.18653/v1/2020.emnlp-tutorials.3}
  {Interpreting predictions of {NLP} models}.
\newblock In \emph{Proceedings of the 2020 Conference on Empirical Methods in
  Natural Language Processing: Tutorial Abstracts}, pages 20--23, Online.
  Association for Computational Linguistics.

\bibitem[{Wang et~al.(2020{\natexlab{a}})Wang, Liu, and
  Song}]{wang2020language}
Chenguang Wang, Xiao Liu, and Dawn Song. 2020{\natexlab{a}}.
\newblock Language models are open knowledge graphs.
\newblock \emph{arXiv preprint arXiv:2010.11967}.

\bibitem[{Wang et~al.(2020{\natexlab{b}})Wang, Tuyls, Wallace, and
  Singh}]{wang-etal-2020-gradient}
Junlin Wang, Jens Tuyls, Eric Wallace, and Sameer Singh. 2020{\natexlab{b}}.
\newblock \href {https://doi.org/10.18653/v1/2020.findings-emnlp.24}
  {Gradient-based analysis of {NLP} models is manipulable}.
\newblock In \emph{Findings of the Association for Computational Linguistics:
  EMNLP 2020}, pages 247--258, Online. Association for Computational
  Linguistics.

\bibitem[{Williams et~al.(2018)Williams, Nangia, and
  Bowman}]{williams2017broad}
Adina Williams, Nikita Nangia, and Samuel Bowman. 2018.
\newblock \href {https://doi.org/10.18653/v1/N18-1101} {A broad-coverage
  challenge corpus for sentence understanding through inference}.
\newblock In \emph{Proceedings of the 2018 Conference of the North {A}merican
  Chapter of the Association for Computational Linguistics: Human Language
  Technologies, Volume 1 (Long Papers)}, pages 1112--1122, New Orleans,
  Louisiana. Association for Computational Linguistics.

\bibitem[{Wolf et~al.(2020)Wolf, Chaumond, Debut, Sanh, Delangue, Moi, Cistac,
  Funtowicz, Davison, Shleifer et~al.}]{wolf2020transformers}
Thomas Wolf, Julien Chaumond, Lysandre Debut, Victor Sanh, Clement Delangue,
  Anthony Moi, Pierric Cistac, Morgan Funtowicz, Joe Davison, Sam Shleifer,
  et~al. 2020.
\newblock Transformers: State-of-the-art natural language processing.
\newblock In \emph{Proceedings of the 2020 Conference on Empirical Methods in
  Natural Language Processing: System Demonstrations}, pages 38--45.

\bibitem[{Xie et~al.(2019)Xie, Lai, Doran, and Kadav}]{Xie2019}
Ning Xie, Farley Lai, Derek Doran, and Asim Kadav. 2019.
\newblock \href {http://arxiv.org/abs/arXiv:1901.06706v1} {{Visual Entailment:
  A Novel Task for Fine-Grained Image Understanding}}.
\newblock \emph{arXiv}, 1901.06706.

\bibitem[{Yatskar et~al.(2016)Yatskar, Zettlemoyer, and
  Farhadi}]{Yatskar_2016_CVPR}
Mark Yatskar, Luke Zettlemoyer, and Ali Farhadi. 2016.
\newblock Situation recognition: Visual semantic role labeling for image
  understanding.
\newblock In \emph{Proceedings of the IEEE Conference on Computer Vision and
  Pattern Recognition (CVPR)}.

\bibitem[{Zhu and Bhat(2020)}]{zhu-bhat-2020-gruen}
Wanzheng Zhu and Suma Bhat. 2020.
\newblock \href {https://doi.org/10.18653/v1/2020.findings-emnlp.9} {{GRUEN}
  for evaluating linguistic quality of generated text}.
\newblock In \emph{Findings of the Association for Computational Linguistics:
  EMNLP 2020}, pages 94--108, Online. Association for Computational
  Linguistics.

\bibitem[{Zhu et~al.(2016)Zhu, Groth, Bernstein, and Fei-Fei}]{zhu2016cvpr}
Yuke Zhu, Oliver Groth, Michael Bernstein, and Li~Fei-Fei. 2016.
\newblock {Visual7W: Grounded Question Answering in Images}.
\newblock In \emph{{IEEE Conference on Computer Vision and Pattern
  Recognition}}.

\end{thebibliography}
\bibliographystyle{acl_natbib}

\clearpage
\appendix

\section{Benchmark creation}\label{app:benchmark}

\subsection{Existence}\label{app:existence}

The {\bf existence} piece has a single instrument and targets instances with \textbf{existential quantifiers}.
Models need to differentiate between examples i) where \textit{there is no entity} of a certain type or ii) where \textit{there is one or more of these entities} visible in an image.

\paragraph{Data sources}
We use the Visual7W visual question answering dataset \citep{zhu2016cvpr} to source examples, starting with the `how many' questions in Visual7W and building a pool of those whose answers are numerals (e.g., 0, 1, 2, etc.).
We use the templates from \citet{parcalabescu2021seeing} to transform question and answer fields into a declarative statement that correctly describes what can be seen in the image, e.g., `Q: How many animals are shown? A: 0' $\rightarrow$ `There are 0 animals shown'.

\paragraph{Foiling method}
Let us use $x =$  `There are N animals shown' as a running example for a correct caption, where $N$ is a number.
If $N>0$, we simply remove $N$ from the sentence, effectively creating the statement $\exists x$ or `There are animals shown'.
If $N=0$, we replace $N$ by `no', creating the statement $\neg\exists x$ or `There are no animals shown'.
If necessary, we fix singular--plural agreement.
To create data with balanced correct and foil classes, we select $50\%$ of our examples from those where the correct answer is originally 0, and the remaining $50\%$ from those where the correct answer is any other number (e.g., 1, 2, etc.).
To create foils, we then simply convert the statement from $\exists x$ to $\neg\exists x$, and vice-versa.

\subsection{Plurality}\label{app:plurality}
The \textbf{plurality} piece has a single instrument, concerned with {\bf semantic number}, that is, the distinction between single entities in an image (`exactly one flower') and multiple instances of the same type (`some flowers'). In this piece, foil candidates are created either by converting a singular NP and its coreferents to a plural, or vice versa. 

\paragraph{Data sources} The data was sourced from the validation split of the COCO 2017 dataset \cite{Chen2015}. Captions are only foiled if their length after tokenization with the pretrained BERT tokenizer\footnote{We use the {\tt bert-large-cased} pretrained tokenizer distributed as part of the {\tt transformers} python library.} is of 80 tokens or less. This is done to minimise the risk that captions and foils need to be truncated to accommodate the input specifications of current pretrained V\&L models.

\paragraph{Foiling method} Foiling is done in two directions: singular-to-plural ({\tt sg2pl}) or plural-to-singular ({\tt pl2sg}). Given a caption, NP chunking is applied to identify all non-pronominal NPs. In the {\tt sg2pl} case, a foiled version of a caption containing a singular NP is created by pluralising the head noun.
We automatically identify anaphoric expressions coreferring to the singular NP within the caption and pluralise them in the same way.
For NPs which are subjects of copular VPs or VPs with an auxiliary requiring subject-verb number agreement (e.g. `N \underline{is} V'), we also pluralise the verb.
Note that this procedure creates a potential foil for every singular NP in the caption; thus, more than one foil candidate can be created for each instance in the source dataset.\footnote{NP chunking is performed using the Spacy v.3 pipeline for English using the {\tt en\_core\_web\_md} pretrained models. Coreference chains are detected using the pretrained English model for Coreferee (\url{github.com/msg-systems/coreferee}). Pluralisation of head nouns is carried out using the {\tt inflect} engine (\url{github.com/jaraco/inflect/}).}
In the {\tt pl2sg} case, the same procedure is carried out, but turning a plural NP, as well as its coreferents, into a singular. We generate all foil candidates using the Checklist framework \cite{ribeiro-etal-2020-beyond}, within which we implement our procedures for data perturbation.

An important consideration, especially in the {\tt pl2sg} case, is that singularising an NP in a foil can still be truth-preserving. Specifically, a caption with a plural NP, such as  `A small copper vase with \underline{some flowers} in it', arguably still entails the version with the singular `(\textellipsis) \underline{a flower}'. As a result, the singular version may still correctly be judged to match the image. One way around this problem is to insert a quantifier in the singular NP which makes it explicit that exactly one instance and no more is intended (e.g. `\underline{exactly one} flower'). This may however result in a biased dataset, with such singular quantifiers acting as signals for singular foils and enabling models to solve the task with no grounding in the visual information. We avoid this by adopting a uniform strategy for both {\tt sg2pl} and {\tt pl2sg}. We determine two singular quantifiers (`exactly one N' and `a single N') and two plural quantifiers (`some N', `a number of N'). When a foil candidate is generated, we alter the {\em original} NP by inserting one of the two quantifiers matching its semantic number, and generate a foil with one of the two quantifiers for the other number.
In the foregoing example, we end up with `A small copper vase with \underline{\gr{some flowers}} / \underline{\red{exactly one flower}} in it.'


After generating all candidate foils, in both directions, we use the GRUEN pretrained model \cite{zhu-bhat-2020-gruen} to score the foils for grammaticality. We only keep foils with a score $\geq 0.8$, and run each foil-caption pair through the NLI model described in Section \ref{subsec:nli}, keeping only pairs whose predicted label is {\em contradiction}, for an initial candidate set of 1000 cases (500 {\tt sg2pl} and 500 {\tt pl2sg}), of which 851 (85.1\%) are considered valid following manual validation (see \S \ref{sec:mturk}).
Figure \ref{fig:foil-caption-distr-part2} shows the distribution of nouns in captions and foils, before and after the validation. Note that the validation process does not result in significant change to the distributions.

\subsection{Counting}\label{app:counting}

The {\bf counting} piece comes in three instruments: {\bf balanced}, {\bf adversarial} and {\bf small numbers}.
All three instruments include instances with \textit{statements about the number of entities visible in an image}.
The model needs to differentiate between examples where \textit{the specific number of entities in the associated image} is correct or incorrect, given the statement.



All three instruments are designed to show whether models learn strategies that generalize beyond the training distribution, and to what extent a model exploits class frequency bias.\footnote{We take the original answer in Visual7W as the example class. E.g., in \textit{There are four zebras}, the class is 4.}
In {\bf counting balanced} we cap the number of examples to a maximum per class and make sure correct/foil classes are balanced, so that models that exploit class frequency bias are penalized.
In {\bf counting adversarial} we make sure that all foils take class $n \in \{0,1,2,3\}$, whereas all correct captions take class $n \in \{n \given n \geq 4\}$. Biased models are expected to favour more frequent classes and these correspond to smaller numbers, therefore models that resort to such biases should perform poorly on this adversarially built test.
Instrument {\bf counting small numbers} is a sanity check where all correct captions and foils have class $n \in \{0,1,2,3\}$, and caption/foil classes are balanced. Models likely have been exposed to many examples in this class set, so with this instrument we assess model performance certain it does not suffer from (class) exposure bias.

\paragraph{Data sources}
We use the Visual7W visual question answering dataset \citep{zhu2016cvpr} and source its `how many' examples, building a pool of those whose answers are numerals (e.g., 0, 1, 2, etc.).
We use the templates from \citet{parcalabescu2021seeing} to transform question and answer fields into a declarative statement that correctly describes what can be seen in the image.


\paragraph{Foiling method}
We create foils by directly replacing the numeral in the correct caption by another numeral. When creating foils we make sure that the class distribution for correct and foiled captions are approximately the same, i.e., there are a similar number of correct and foiled examples in each class in each instrument.
The only exception is the counting adversarial instrument, where the classes used in correct and foiled captions are disjoint, i.e., $n \in \{0,1,2,3\}$ and $n \in \{n \given n \geq 4\}$, respectively.
See Figure \ref{fig:foil-caption-distr-part1} for a visualisation of these distributions.


\subsection{Spatial relations}\label{app:relations}
The \textbf{relations} piece has one instrument and focuses on the ability of models to distinguish between different spatial relations, as expressed by prepositions. Foils therefore consist of captions identical to the original except for the replacement of a spatial preposition.

\paragraph{Data sources} Data
was sourced from the COCO 2017 validation split \cite{Chen2015}. To generate foil candidates, we first extracted from the original COCO captions all the sequences consisting of one or more consecutive prepositions (e.g., `on' or `out of'). Foils are generated by detecting these preposition spans, and replacing them with another preposition span attested in the list.

\paragraph{Foiling method} To generate foils, we mask the preposition span in an original caption, and use SpanBERT \cite{joshi-etal-2020-spanbert}, a pretraining method based on BERT \cite{devlin2019bert}.\footnote{We use SpanBERT with the pretrained {\tt bert-large-cased} model distributed as part of the {\tt transformers} Python library. } The advantage of SpanBERT over BERT is that in a masked language modelling context, with masks spanning more than a single word, SpanBERT predicts sequences and takes into account their joint probability, whereas BERT trained with standard Masked Language Modelling can only predict single tokens independently. With SpanBERT, we generate replacements of between 1 and 3 tokens in length, in each case retaining only the best prediction out of the top $k$ which matches one of the preposition sequences in the pre-extracted list. 


After all candidates are generated, we apply GRUEN \cite{zhu-bhat-2020-gruen} to score the foils for grammaticality, and further apply the NLI model descibed in Section \ref{subsec:nli} to label the entailment relationship between caption and foil pairs. From the resulting data, we sample as follows: i) we keep only caption-foil pairs labelled as {\em contradiction}, where the GRUEN grammaticality score is $\geq 0.8$; ii) for every caption-foil pair sampled where $p$ is replaced with $q$, we search for another caption-foil pair where $q$ is replaced with $p$, if present. This strategy yields a roughly balanced dataset, where no single preposition or preposition sequence is over-represented in captions or foils. 

These processes result in an initial set of 614 cases, of which 535 (87.1\%) are selected following manual validation described in \S \ref{sec:mturk}. 

Figure \ref{fig:foil-caption-distr-part1} shows proportions in captions and foils of the prepositions. E.g.: `A cat plays with a pocket knife \gr{\underline{on}} / \red{\underline{underneath}} a table.'

As with plurals, we implement procedures for foil candidate generation by extending the \texttt{perturb} functionality in Checklist \cite{ribeiro-etal-2020-beyond}.

\subsection{Actions}\label{app:actions}
The \textbf{action} piece consists of two instruments: i) \textbf{action replacement} and ii) \textbf{actant swap}.
They are testing a V\&L model's capability of i) identifying whether an \textit{action} mentioned in the textual modality 
matches the action seen in the image or not 
(e.g. `a man \gr{\underline{shouts}} / \red{\underline{smiles}} at a woman') 
and ii) correctly identifying the \textit{participants} of an action and the \textit{roles} they are playing in it 
(e.g., given the picture in Table \ref{tab:phenomena}: is it the man or the woman who shouts?).


\paragraph{Data source}
For creating interesting foils with \emph{diverse} actions, we focus on the SWiG dataset \cite{DBLP:conf/eccv/PrattYWFK20} that comprises 504 action verbs annotated with semantic roles and their fillers, which are grounded in images of the \textit{imSitu} dataset \cite{Yatskar_2016_CVPR}. We generate English captions for the images 
using SimpleNLG \cite{gatt-reiter-2009-simplenlg}\footnote{SimpleNLG is a surface realization engine that -- given some content and crucial syntactic specifications -- performs surface generation including  morphological adjustments.}. For generation we use the specified \textit{action verb}, the  realized FrameNet semantic roles and their annotated filler categories (see Table \ref{tab:phenomena} for \textit{shout}: \textsc{Agent}: man, \textsc{Addressee}: woman), and generate short captions, with realization of  two roles in active form.
We apply various filters to ensure high quality of the generated captions using diverse metrics\footnote{We use the GRUEN metric \citep{zhu-bhat-2020-gruen} that scores grammaticality, naturalness and coherence of generations and compute perplexity with GPT-2 to rank alternative outputs. We determined appropriate  thresholds based on manual judgements of acceptability and chose the highest-ranked candidates. The final data quality is controlled by crowdsourced annotation with AMT.} and manual checks through AMT crowdsourcing.

\paragraph{Foiling method}
When creating the \textbf{action replacement} instrument, we need to make sure that the action replacement suits the context. We propose action replacements with BERT \cite{devlin2019bert} that need to satisfy three conditions: 1) the proposed action verbs originate from the SWiG dataset -- otherwise new verbs are introduced on the foil side only, which may induce biases;
2) the frequency distribution of action verbs on the caption and on the foil side is approximately the same (cf. Figure \ref{fig:foil-caption-distr-part2});
3) we constrain the replacement verbs to be either antonyms of the original verb or at least not synonyms, hyponyms or hypernyms to the original, according to WordNet \cite{wordnet} in order to avoid situations where replacements are almost synonymous to the original action.
The \textbf{actant swap} instrument is based on the original image annotations, but swaps the two role fillers (e.g., `A woman shouts at the man.' for the image in Table \ref{tab:phenomena}). To avoid agreement mistakes, we \textit{generate} these foils using the inverted role fillers as input. 

We plot caption and foil word frequency distributions for action replacement in Figure \ref{fig:foil-caption-distr-part2}. We do not plot statistics for the actant swap instrument since by construction it cannot suffer from distributional bias since caption and foil contain the same words up to a \emph{permutation}.

\subsection{Coreference}\label{app:coref}

The \textbf{coreference} piece consists of two pieces: \textbf{coreference standard} and \textbf{coreference clean}. It aims to uncover whether V\&L models are able to perform pronoun coreference resolution. The coreference phenomenon encompasses both cases where i) the pronoun refers to a noun (phrase) and both the pronoun and the (noun) phrase are grounded in the visual modality (e.g. `\underline{A woman} is driving a motorcycle. Is \underline{she} wearing a helmet?'), and cases where ii) the pronoun refers directly to a region in the image or even to the whole image (e.g. `A man is sitting on a bench. Is \underline{this} outside?').

\begin{figure}
    \centering
    \includegraphics[width=.9\linewidth]{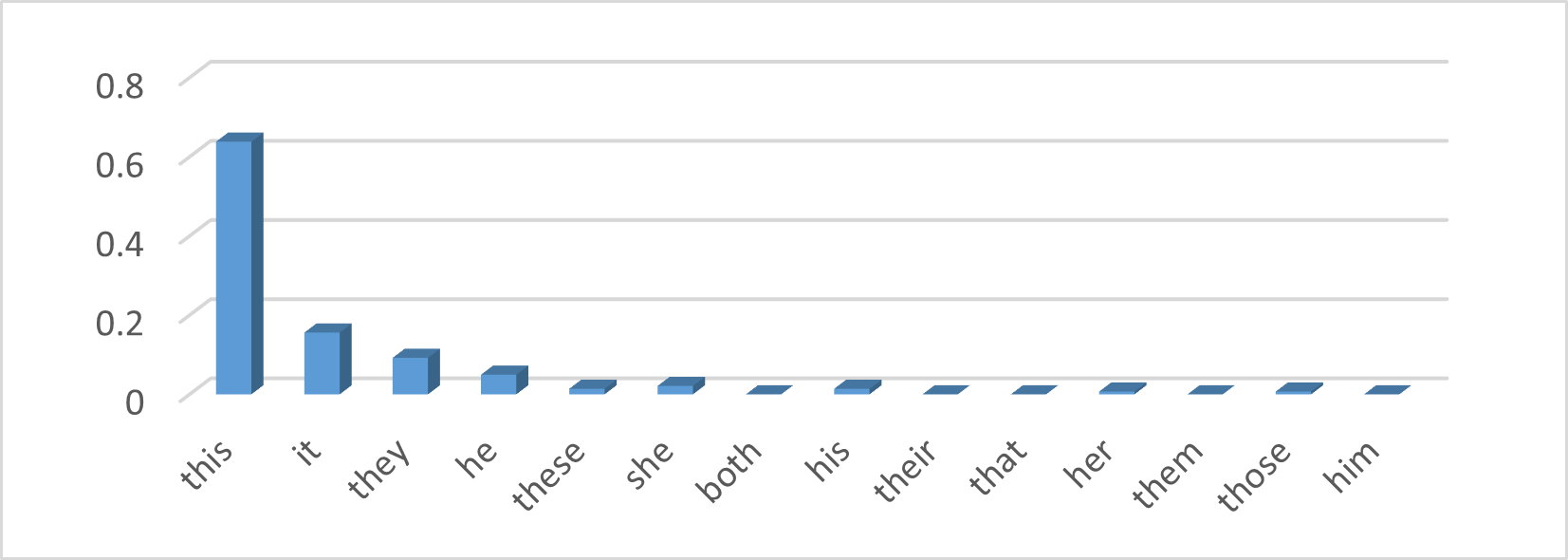}
    \caption{Normalized pronoun frequencies in the coreference subset.}
    \label{fig:coref-pron-freqs}
\end{figure}

\paragraph{Data source}
We source the data from VisDial v1.0 \cite{visdial}, which contains images from MSCOCO \cite{Lin-etal:2014:mscoco}, their captions and dialogues about the images in form of Q\&A sequences.
To ensure that the coreference phenomenon is present in the [\emph{Caption. Question? Yes/No.}] formulations, we check whether pronouns are present in the \emph{question}. The list of pronouns and their frequencies in our train-val-test splits are represented in Figure \ref{fig:coref-pron-freqs}.

The \textbf{coreference standard} instrument contains 916 data samples (708 are valid\footnote{The majority of manual annotators validated that the caption describes the image but the foil does not.}) from the VisDial's training set. The data of \textbf{coreference clean} instrument consisting of 141 samples (104 are valid), originates from VisDial's validation set. With models that have been trained on VisDial, we would be in the situation where models are tested on their training data. Therefore we also have the \emph{coreference clean instrument} based on the validation set of VisDial to test models safely. Unfortunately, we cannot use VisDial's test set because the required question-answers annotations necessary for foiling are withheld.

\paragraph{Foiling method}
When foiling, we take the image description of the form [\emph{Caption. Question? Yes/No.}] and exchange the answer: \emph{yes} \textrightarrow \emph{no} and vice-versa (see example in Table \ref{tab:phenomena}). This way, we keep the full textual description including pronoun and noun (phrase) intact, hence ensuring that the coreference phenomenon is present and valid in the foil too, and rely on the model to interpret affirmation and negation correctly. Note that we rely on the capability of models to correctly interpret negation also in the existence piece (cf.\ \S \ref{subsec:existence}).

Arguably, coreference is the most difficult phenomenon to foil in \dataset{}. Especially in cases where pronouns refer to a noun (phrase) (e.g., `\underline{A woman} is driving a motorcycle. Is \underline{she} wearing a helmet? Yes.'), exchanging the pronoun with another pronoun would generate incoherent and unlikely sequences\footnote{Even more, the possibilities of exchanging pronouns with pronouns in grammatical ways are very limited: \emph{she -- he} but not \emph{she -- they /  her / their}.} (e.g., `\underline{A woman} is driving a motorcycle. Is \underline{he} wearing a helmet?'), and exchanging it with a noun phrase would furthermore break the pronoun coreference phenomenon because there would be no pronoun anymore (e.g., `\underline{A woman} is driving a motorcycle. Is \underline{the man} wearing a helmet?'). Therefore when foiling the coreference piece, we aim to keep the original description intact for ensuring the preservation of the coreference phenomenon. Hence we rely on the answers containing \emph{yes} or \emph{no}\footnote{If the answer is longer than just \emph{yes/no} (e.g., `Yes, she is') we shorten it to \emph{yes/no}.} and exchange affirmative to negative answers and vice-versa.

\subsection{FOIL it! data}\label{app:foilit}
We include an additional piece in \dataset{} consisting of 1000 randomly sampled entries from the \textit{FOIL it!} dataset \citep{shekhar-etal-2017-foil}.
Each entry in \textit{FOIL it!} consists of an MSCOCO \citep{Lin-etal:2014:mscoco} image and a foiled caption where a noun phrase depicting an object visible in the image was replaced by a semantically related noun phrase.
Since examples in the \textit{FOIL it!} dataset are linked to MSCOCO, we use these links to retrieve one correct caption from the five captions available for the image, and create an image--caption--foil triple.
From the original 1000 entries, 943 have been validated by our manual annotation procedure (in Appendix \ref{app:mturk}). 
Please refer to \citet{shekhar-etal-2017-foil} for more details.

\section{Evaluation metrics}\label{app:metrics}
We evaluate pretrained V\&L models on \dataset{} using {\bf accuracy} ($acc$), the overall accuracy on all classes; {\bf precision} or \textit{positive predictive value} ($p_c$), which measures the proportion of correctly identified \textit{correct captions}; and {\bf foil precision} or \textit{negative predictive value} ($p_f$), which measures the proportion of correctly identified \textit{foiled} examples;
{\bf pairwise ranking accuracy} $acc_r$, computed using the image-sentence alignment score $\phi$ that the model assigns to correct and foiled image-text pairs;
and {\bf area under the receiver operating characteristic curve} (AUROC)---a classic metric used in machine learning classification problems---which in our case measures how well models distinguish correct vs. foiled examples across different prediction thresholds.
The AUROC has a probabilistic interpretation and can be understood as the probability that a model will assign a higher score to a randomly chosen correct example relative to a randomly chosen foil.


\newcolumntype{R}[1]{>{\raggedleft\let\newline\\\arraybackslash\hspace{0pt}}m{#1}}

\begin{table*}[ht]
  \small
  \centering
  \resizebox{\linewidth}{!}{
  \begin{tabular}{@{}%
  R{.1\linewidth}%
  M{.17\linewidth}M{.19\linewidth}M{.17\linewidth}M{.17\linewidth}M{.15\linewidth}@{}}
    \toprule
     & {\bf CLIP} & {\bf LXMERT} & {\bf ViLBERT} & {\bf ViLBERT 12-in-1} & {\bf VisualBERT} \\
     & \citep{radford2021learning} & \citep{tan-bansal-2019-lxmert} & \citep{lu2019vilbert} & \citep{lu2020vilbert12in1} & \citep{li2019visualbert} \\
    \midrule
    model type   & separate image and text encoders & dual stream & dual stream & dual stream & single stream\\
    \midrule
    pretraining data   & 400M image-text pairs & MSCOCO & Conceptual Captions & Conceptual Captions & MSCOCO \\
    \midrule
    pretraining tasks & ISA & ISA, MLM, MOP, VQA & ISA, MLM, MOP & ISA, MLM, MOP & ISA, MLM, MOP \\
    \midrule
    finetuning   & -- & VQA & -- & 12 V\&L tasks & -- \\
    \bottomrule
  \end{tabular}
  }
  \caption{V\&L models evaluated with \dataset{} in our experiments. {\bf ISA}: image-sentence alignment; {\bf MLM}: masked language modelling; {\bf MOP}: masked object prediction; {\bf VQA}: visual question answering.}
  \label{tab:models}
\end{table*}

\begin{table*}[t!]
    \small
    \centering
    \resizebox{\linewidth}{!}{
    \begin{tabular}{ l r r@{\hskip 0.2in}r@{\hskip 0.2in}rrr@{\hskip 0.2in}r@{\hskip 0.2in}rr@{\hskip 0.2in}rr@{\hskip 0.2in}r@{\hskip 0.2in}r }
        \toprule
        \multirow{2}{*}{\bf Metric} & \multirow{2}{*}{\bf Model} &
        \multicolumn{1}{c|}{\bf Existence} & \multicolumn{1}{c|}{\bf Plurality} & \multicolumn{3}{c|}{\bf Counting} & \multicolumn{1}{c|}{\bf Sp.rel.$\ddagger$} & \multicolumn{2}{c|}{\bf Action} & \multicolumn{2}{c|}{\bf Coreference} & \multicolumn{1}{c|}{\multirow{2}{*}{\bf Foil-it!}} & \multicolumn{1}{c}{\multirow{2}{*}{\bf Avg.}} \\
        && \multicolumn{1}{c|}{quantifiers} & \multicolumn{1}{c|}{number} & \multicolumn{1}{c}{balanced} & \multicolumn{1}{c}{sns.$\dagger$} & \multicolumn{1}{c|}{adv.$\dagger$} & \multicolumn{1}{c|}{relations} & \multicolumn{1}{c}{repl.$\dagger$} & \multicolumn{1}{c|}{actant swap} & \multicolumn{1}{c}{standard} & \multicolumn{1}{c|}{clean} &
        \multicolumn{1}{c|}{}\\ 
        \midrule
        & \multicolumn{1}{r}{Random} & 50.0 & 50.0 & 50.0 & 50.0 & 50.0 & 50.0 & 50.0 & 50.0 & 50.0 & 50.0 & 50.0 & 50.0 \\
        \midrule
        \multirow{7}{*}{$acc_r$}
        & \multirow{1}{*}{GPT1$^*$} & 61.8 & 53.1 & 51.2 & \redtable{48.7} & 69.5 & {\bf 77.2} & 65.4 & 72.2 & \redtable{45.6} & \redtable{45.2} & 77.5 & 60.7 \\
        & \multirow{1}{*}{GPT2$^*$} & 58.0 & 51.9 & 51.6 & \redtable{49.8} & \redtable{45.3} & 75.0 & 66.8 & {\bf 76.9} & 54.5 & \redtable{50.0} & 80.7 & 60.1 \\
        & \multirow{1}{*}{CLIP} & 66.9 & 56.2 & 62.1 & 62.5 & 57.5 & 64.3 & {\bf 75.6} & 68.6 & 52.1 & \redtable{49.7} & {\bf 88.8} & 64.0 \\
        & \multirow{1}{*}{LXMERT} & 78.6 & 64.4 & 62.2 & 69.2 & \redtable{42.6} & 60.2 & 54.8 & \redtable{45.8} & \redtable{46.8} & \redtable{44.2} & 87.1 & 59.6 \\
        & \multirow{1}{*}{ViLBERT} & 65.5 & 61.2 & 58.6 & 62.9 & 73.7 & 57.2 & 70.7 & 68.3 & \redtable{47.2} & \redtable{48.1} & 86.9 & 63.7 \\
        & \multirow{1}{*}{12-in-1} & {\bf 95.6} & {\bf 72.4} & {\bf 76.7} & {\bf 80.2} & {\bf 77.3} & 67.7 & 65.9 & 58.9 & {\bf 75.7} & {\bf 69.2} & 86.9 & {\bf 75.1} \\
        & \multirow{1}{*}{VisualBERT} & \redtable{39.7} & \redtable{45.7} & \redtable{48.2} & \redtable{48.2} & \redtable{50.0} & \redtable{39.7} & \redtable{49.2} & \redtable{44.4} & \redtable{49.5} & \redtable{47.6} & \redtable{48.5} & \redtable{46.4}\\
        \cmidrule{2-14}
        \multirow{4}{*}{$acc$}
        & \multirow{1}{*}{LXMERT} & 55.8 & 55.1 & 52.0 & 55.4 & \redtable{49.9} & 50.8 & 51.1 & \redtable{48.5} & \redtable{49.8} & \redtable{49.0} & 70.8 & 53.5 \\
        & \multirow{1}{*}{ViLBERT} & \redtable{2.4} & 50.3 & 50.7 & 50.6 & 51.8 & \redtable{49.9} & 52.6 & 50.4 & \redtable{50.0} & \redtable{50.0} & 55.9 & 51.3 \\
        & \multirow{1}{*}{12-in-1} & {\bf 89.0} & {\bf 62.0} & {\bf 64.9} & {\bf 69.2} & {\bf 66.7} & {\bf 53.4} & {\bf 57.3} & {\bf 52.2} & {\bf 54.4} & {\bf 54.3} & {\bf 71.5} & {\bf 63.2} \\
        & \multirow{1}{*}{VisualBERT} & \redtable{49.3} & \redtable{46.5} & \redtable{48.3} & \redtable{47.8} & \redtable{50.0} & \redtable{49.3} & \redtable{48.8} & \redtable{49.7} & \redtable{50.0} & \redtable{50.0} & \redtable{46.6} & \redtable{48.8} \\
        \cmidrule{2-14}
        \cmidrule{2-14}
        \multirow{4}{*}{$p_c$}
        & \multirow{1}{*}{LXMERT} & \redtable{41.6}  & 68.0 & 50.9 & \redtable{50.0} & 61.5 & 73.1 & \redtable{35.8} & \redtable{36.8} & 81.2 & 80.8 & 72.3 & 59.3 \\
        & \multirow{1}{*}{ViLBERT} & 56.8 & {\bf 98.5} & {\bf 77.0} & 76.6 & {\bf 86.1} & {\bf 98.3} & {\bf 93.2} & {\bf 93.7} & {\bf 98.7} & {\bf 98.1} & {\bf 98.8} & {\bf 88.7} \\
        & \multirow{1}{*}{12-in-1} & {\bf 85.0} & 90.7 & 64.3 & {\bf 76.7} & 59.5 & 93.5 & 66.7 & 66.8 & 92.9 & 95.2 & 94.3 & 80.5 \\
        & \multirow{1}{*}{VisualBERT} & \redtable{1.3}  & \redtable{0.3} & \redtable{0.0} & \redtable{0.0} & \redtable{0.0} & \redtable{1.3} & \redtable{0.0} & \redtable{0.0} & \redtable{0.0} & \redtable{0.0} & \redtable{0.2} & \redtable{0.3} \\
        \cmidrule{2-14}
        \multirow{4}{*}{$p_f$}
        & \multirow{1}{*}{LXMERT} & 70.1 & \redtable{42.2} & 53.0 & 60.8 & \redtable{37.3} & \redtable{28.4} & 66.4 & 60.2 & \redtable{18.4} & \redtable{17.3} & 69.3 & \redtable{47.6} \\
        & \multirow{1}{*}{ViLBERT} & \redtable{47.9} & \redtable{2.1} & \redtable{24.4} & \redtable{24.7} & \redtable{17.5} & \redtable{1.5} & \redtable{11.9} & \redtable{7.1} & \redtable{1.3} & \redtable{1.9} & \redtable{12.9} & 13.9 \\
        & \multirow{1}{*}{12-in-1} & 93.1 & \redtable{33.4} & 65.6 & 61.7 & 74.0 & \redtable{13.3} & \redtable{47.8} & \redtable{37.6} & \redtable{15.8} & \redtable{13.5} & \redtable{48.8} & \redtable{45.9} \\
        & \multirow{1}{*}{VisualBERT} & \bf{97.3} & \bf{92.8} & \bf{96.7} & \bf{95.7} & \bf{100.0}  & \bf{97.3} & \bf{97.6} & \bf{99.4} & \bf{100.0} & \bf{100.0} & \bf{93.0} & \bf{97.3} \\
        \cmidrule{2-14}
        \multirow{4}{*}{min($p_c$, $p_f$)}
        & \multirow{1}{*}{LXMERT} & \redtable{41.6}  & \redtable{\bf 42.2} & 50.9 & \redtable{50.0} & \redtable{37.3} & \redtable{\bf 28.4} & \redtable{35.8} & \redtable{36.8} & \redtable{\bf 18.4} & \redtable{\bf 17.3} & {\bf 69.3} & \redtable{38.9} \\
        & \multirow{1}{*}{ViLBERT} & \redtable{47.9} & \redtable{2.1} & \redtable{24.4} & \redtable{24.7} & \redtable{17.5} & \redtable{1.5} & \redtable{11.9} & \redtable{7.1} & \redtable{1.3} & \redtable{1.9} & \redtable{12.9} & \redtable{13.9} \\
        & \multirow{1}{*}{12-in-1} & {\bf 85.0} & \redtable{33.4} & {\bf 64.3} & {\bf 61.7} & {\bf 59.5} & \redtable{13.3} & \redtable{\bf 47.8} & \redtable{\bf 37.6} & \redtable{15.8} & \redtable{13.5} & \redtable{48.8} & \redtable{\bf 43.7} \\
        & \multirow{1}{*}{VisualBERT} & \redtable{1.3}  & \redtable{0.3} & \redtable{0.0} & \redtable{0.0} & \redtable{0.0} & \redtable{1.3} & \redtable{0.0} & \redtable{0.0} & \redtable{0.0} & \redtable{0.0} & \redtable{0.2} & \redtable{0.3} \\
        \cmidrule{2-14}
        \multirow{4}{1cm}{$AUROC$ $\times 100$}
        & \multirow{1}{*}{LXMERT} & 60.5 & 57.3 & 53.8 & 57.7 & 50.5 & 51.9 & 52.1 & \redtable{47.6} & \redtable{49.8} & \redtable{49.5} & 76.9 & 55.2 \\
        & \multirow{1}{*}{ViLBERT} & 52.5 & 54.1 & 50.8 & 51.6 & 53.5 & 51.2 & 57.2 & {\bf 57.8} & \redtable{49.9} & \redtable{49.9} & 75.2 &  54.9 \\
        & \multirow{1}{*}{12-in-1} & {\bf 96.3} & {\bf 67.4} & {\bf 72.0} & {\bf 77.8} & {\bf 75.1} & {\bf 55.8} & {\bf 61.3} & 55.0 & {\bf 59.8} & {\bf 59.6} & {\bf 81.0} & {\bf 69.2} \\
        & \multirow{1}{*}{VisualBERT} & \redtable{28.9} & \redtable{29.0} & \redtable{24.5} & \redtable{16.5} & \redtable{20.9} & \redtable{45.2} & \redtable{17.7} & \redtable{36.3} & \redtable{45.3} & \redtable{46.3} & \redtable{28.5} & \redtable{30.8} \\
        \bottomrule
    \end{tabular}
    }
    \caption{Performance of unimodal and multimodal models on the \dataset{} benchmark according to different metrics. We bold-face the best overall result per metric, and underscore all results below (or at) the random baseline.
    $acc_r$ is a pairwise ranking accuracy where a prediction is considered correct if $p(caption,img) > p(foil,img)$.
    Precision $p_c$ and foil precision $p_f$ are \emph{competing} metrics where na\"{i}vely increasing one can decrease the other: therefore \emph{looking at the smaller number among the two gives a good intuition of how informed is a model prediction}.
    $\dagger${\bf sns.} Counting small numbers. {\bf adv.} Counting adversarial. {\bf repl.} Action replacement. $\ddagger$ {\bf Sp.rel.} Spatial relations. $^*$Unimodal text-only models that do not use images as input. CLIP is only tested in pairwise ranking mode (fn.\ 6).
    }
    \label{tab:results_all}
\end{table*}

With $acc_r$, a prediction is considered successful, if given an image ($i$) paired with a correct ($c$) versus a foil ($f$) text, the score of the positive/correct pair is greater than that of the foiled pair.
\begin{equation*} \label{eq:metric_pairwise}
    \begin{split}
       acc_r &= \frac{\sum_{(i,c) \in C} \sum_{f \in F} s(i,c,f)}{|C| + |F|}, \\
       s(i,c,f) &=
       \begin{cases}
           1, & \text{if } \phi(i, f) \le \phi(i, c),\\
           0, & \text{otherwise,}
       \end{cases}
    \end{split}
\end{equation*}
\noindent
where $C$ is the set of correct image-caption pairs ($i,c$), and $F$ is the set of foils for the pair ($i,c$).

The \textbf{pairwise accuracy} $acc_r$ is important for two reasons: First, it enables V\&L models to be evaluated on \dataset{} without a binary classification head for classifying image-sentence pairs as correct or foiled. For example, CLIP \cite{radford2021learning} is a model that computes a score given an image-sentence pair. This score can be used to compare the scores of a correct image-sentence pair and the corresponding foiled pair.
By contrast, a model like LXMERT \cite{tan-bansal-2019-lxmert} has a binary image-sentence classification head and can predict a correct pair independently of the foiled pair (and vice-versa). Second, $acc_r$ enables the evaluation of unimodal models on \dataset{}, as motivated 
in \S \ref{subsec:why-unimodal}.
In Table \ref{tab:results_all}, we show results for all models investigated according to all above-mentioned metrics.

\begin{table*}[!t]
    \small
    \centering
    \resizebox{\linewidth}{!}{
    \begin{tabular}{llrrrccccc}
        \toprule
        {\bf Piece} & {\bf Instrument} & {\bf \#Inst.} & {\bf \#Valid (\%)} & {\bf \#Unan. (\%)} & {\bf \#Lex.it.} & {\bf JS} & {\bf JS Val.} & { $\bm{\alpha}$ } & { \bf $\bm{\alpha}$ Valid} \\
        \midrule
        {\bf Existence} & {\em Existential quantifiers} & 534 & 505 (94.6) & 410 (76.8) & 25 & 0.628 & 0.629 & 0.607 & 0.644 \\
        \cmidrule{2-10}
        {\bf Plurality} & {\em Semantic Number} & 1000 & 851 (85.1) & 617 (61.7) & 704 & 0.742 & 0.766 & 0.303 & 0.359  \\
        \cmidrule{2-10}
        \multirow{3}{*}{\bf Counting} & {\em Balanced} & 1000 & 868 (86.8) & 598 (59.8) & 25 & 0.070 & 0.082 & 0.361 & 0.423 \\
                        &  {\em Small numbers} & 1000 & 900 (90.0) & 637 (63.7) & 4 & 0.059 & 0.071 & 0.417 & 0.473 \\
                        & {\em Adversarial} & 756 & 691 (91.4) & 522 (69.0) & 27 & 1.000 & 1.000 & 0.387 & 0.441 \\
        \cmidrule{2-10}
        {\bf Relations} & {\em Prepositions} & 614 & 535 (87.1) & 321 (52.3) & 38 & 0.083 & 0.114 & 0.210 & 0.229\\
        \cmidrule{2-10}
        \multirow{2}{*}{\bf Actions} & {\em Replacement} & 779 & 648 (83.2) & 428 (54.9) & 262 & 0.437 & 0.471 & 0.229 & 0.318  \\
                        & {\em Actant swap} & 1042 & 949 (91.1) & 756 (72.6) & 467 & 0.000 & 0.000 & 0.386 &  0.427 \\
        \cmidrule{2-10}
        \multirow{2}{*}{\bf Coreference} & {\em standard: VisDial train} & 916 & 708 (77.3) & 499 (54.5) & 2 & 0.053 & 0.084  & 0.291 & 0.360 \\
                        & {\em clean: VisDial val} & 141 & 104 (73.8) & 69 (48.9) & 2 & 0.126 & 0.081 & 0.248 & 0.375 \\
         \cmidrule{2-10}
         {\bf Foil-It!} & {\em noun replacement} & 1000 & 943 (94.3) & 811 (81.1) & 73 & 0.426 & 0.425 & 0.532 & 0.588 \\
         \cmidrule{2-10}
         {\bf Overall} & & 8782 & 7702 (87.7) & 5668 (73.6) \\
         \bottomrule
    \end{tabular}
    }
    \caption{Manual validation results for each piece in \dataset{}, as well as for the Foil-it dataset. {\em \#Inst.}: number of instances for linguistic phenomenon. {\em \#Valid (\%)}: number (percent) of cases for which at least 2 out of 3 annotators chose the caption; {\em \#Unan. (\%)}: number (percent) of cases for which all annotators chose the caption; {\em \#Lex.It.}:
    number of phrases or lexical items in the vocabulary that differ between foils and captions; {\em JS}: Jensen-Shannon divergence between foil-caption distributions for all instances in the whole instrument; {\em JS Val.}: Jensen-Shannon divergence between foil-caption distribution for the valid subset of the instrument, after sub-sampling; {\em $\alpha$}: Krippendorff's $\alpha$ coefficient computed over all the instances; {\em $\alpha$ valid}: Krippendorff's $\alpha$ coefficient computed over the {\em Valid} instances.}
    \label{tab:mturk-results}
\end{table*}


\section{Filtering methods}\label{app:filtering}

\paragraph{NLI filtering} For NLI filtering we make use of the \textit{HuggingFace} \cite{wolf2020transformers} implementation of ALBERT (xxlarge-v2) that was already finetuned on the concatenation of SNLI \cite{bowman2015large}, MultiNLI \cite{williams2017broad}, FEVER-NLI \cite{nie2019combining} and ANLI datasets \cite{nie2019adversarial}. The model is the best performing on the ANLI benchmark leaderboard\footnote{\url{github.com/facebookresearch/anli}} and it achieves 90\% accuracy on MultiNLI devset.

\section{Vision \& Language and Unimodal Models}\label{app:models}

In Table \ref{tab:models} we summarise the five V\&L models used in our experiments, their architecture, pretraining tasks and data, and fine-tuning tasks (if any).

\paragraph{CLIP}
CLIP \cite{radford2021learning} is composed of two transformer-based text and an image encoders. These are jointly trained on 400M image-text pairs through contrastive learning for predicting high scores for paired image-text examples and low scores when image-text samples are not paired in the dataset. CLIP has shown zero-shot capabilities in e.g. object classification, OCR, activity recognition \cite{radford2021learning}. \citet{goh2021multimodal} have shown the existence of multimodal neurons in CLIP, responding to the same topic regardless of whether it is represented in an image, drawing or handwritten text. We use CLIP's image-text alignment scores for benchmarking on \dataset{}: Given an image, we compare whether CLIP\footnote{\url{github.com/openai/CLIP}} 
predicts higher image-text similarity for the correct or for the foiled caption.

\paragraph{LXMERT}
LXMERT \cite{tan-bansal-2019-lxmert} is a dual-stream transformer model combining V\&L through cross-modal layers. It is pretrained on MSCOCO \cite{Lin-etal:2014:mscoco} and on multiple VQA datasets for (i) multimodal masked word and object prediction, (ii) image-sentence alignment, i.e., determining whether a text corresponds to an image or not, and (iii) question-answering. For benchmarking on \dataset{}, we use LXMERT's\footnote{\url{github.com/huggingface/transformers}} 
image-sentence alignment head.

\paragraph{ViLBERT and ViLBERT 12-in-1} ViLBERT \citep{lu2019vilbert} is a BERT-based transformer architecture that combines V\&L on two separate streams by co-attention layers. It is pretrained on Google Conceptual Captions \citep{sharma2018conceptual} on (i) multimodal masked word and object prediction; and (ii) image-sentence alignment.
ViLBERT 12-in-1 \citep{lu2020vilbert12in1} further finetuned a ViLBERT model checkpoint on 12 different tasks including VQA, image retrieval, phrase grounding and others.\footnote{\url{github.com/facebookresearch/vilbert-multi-task}}
We use the image-sentence alignment head of the publicly available model checkpoints for ViLBERT\footnote{\url{https://dl.fbaipublicfiles.com/vilbert-multi-task/pretrained_model.bin}} and ViLBERT 12-in-1\footnote{\url{https://dl.fbaipublicfiles.com/vilbert-multi-task/multi_task_model.bin}}.

\paragraph{VisualBERT} VisualBERT \cite{li2019visualbert} is also a BERT-based transformer. Its single-stream architecture encodes image regions and linguistic features via a transformer stack, using self-attention to discover the alignments between the two modalities. VisualBERT is pretrained on MSCOCO captions \cite{Chen2015} on two tasks: (i) masked language modelling, and (ii) sentence-image prediction. The latter is framed as an extension of the next sentence prediction task used with BERT. Inputs consist of an image and a caption, with a second caption which has a 50\% probability of being random. The goal is to determine if the second caption is also aligned to the image.
In our experiments, we use the publicly available implementation of VisualBERT\footnote{\url{github.com/uclanlp/visualbert}}.

\paragraph{GPT-1 and GPT-2 -- Unimodal models}
GPT1 \cite{radford2018improving} and GPT2 \cite{radford2019language} are transformer-based autoregressive language models pretrained on English data through self-supervision. We test whether our benchmark is solvable by these unimodal models by computing the perplexity of the correct sentence and compare it to the perplexity of the foiled sentence. In case the computed perplexity is higher for the foil than for the correct sentence,
we assume that the correctly detected foiled caption may possibly suffer
from a \textbf{plausibility bias} (as described in section \ref{subsec:why-unimodal}) or from other biases (e.g. a model's preference towards affirmative or negative sentences).

\begin{figure*}[t]
    \centering
    \includegraphics[width=\linewidth]{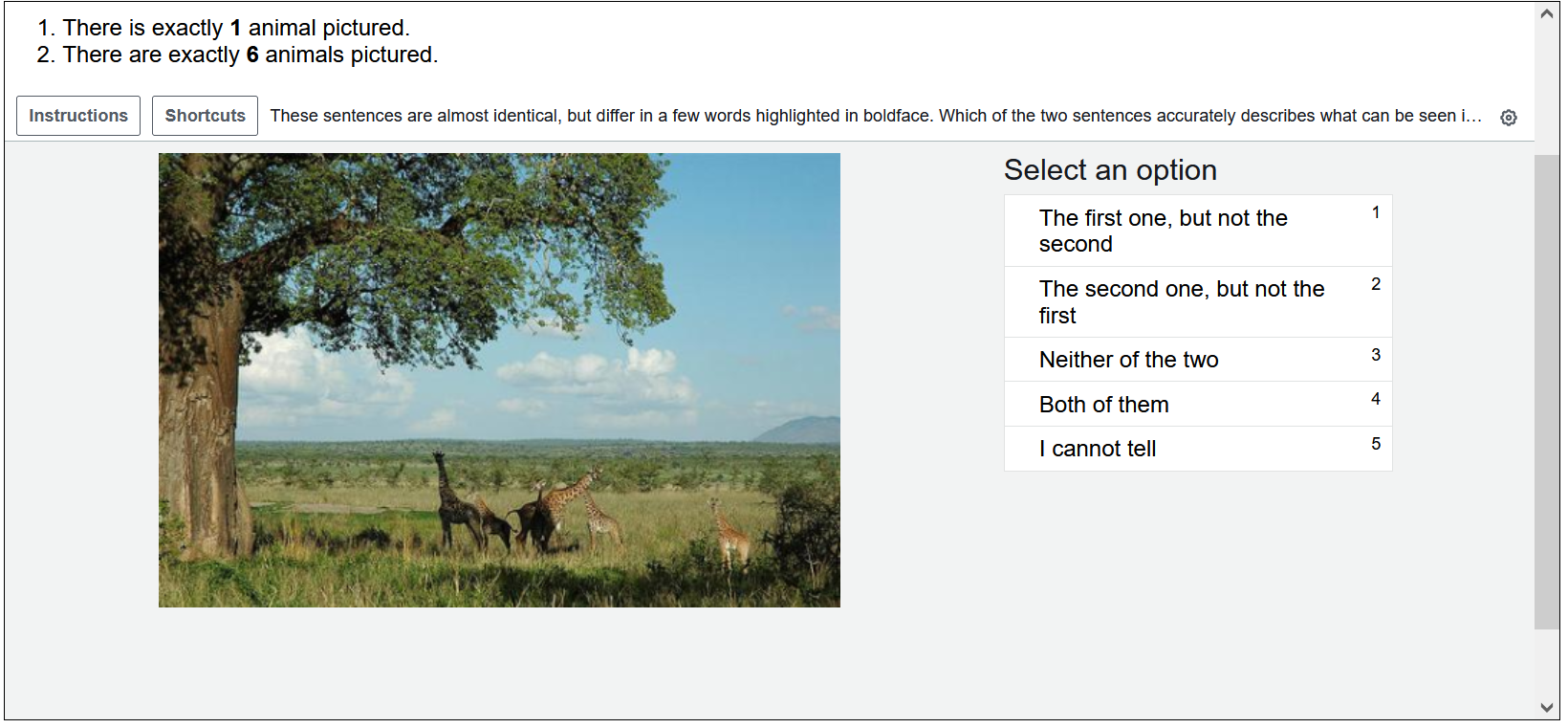}
    \caption{Example of an instance from the validation study. The example is from the counting piece, {\em adversarial} instrument (see Section \ref{subsec:counting}).}
    \label{fig:mturk-example}
\end{figure*}

\section{Mechanical Turk Annotation and Evaluation}\label{app:mturk}
\paragraph{Setup} 
The validation study was conducted on all the data for each instrument in \dataset{}, as well as for the FOIL it! data \cite{shekhar-etal-2019-beyond}. Each instance consisted of an image, a caption and a foiled version of the caption, as shown in Figure \ref{fig:mturk-example}. Annotators received the following general instructions:

\begin{quote}
    \emph{
    You will see a series of images, each accompanied by two short texts. Your task is to judge which of the two texts accurately describes what can be seen in the image. }
\end{quote}

Each instance was accompanied by the caption and the foil, with the ordering balanced so that the caption appeared first 50\% of the time. In each instance, the caption and foil were placed above each other, with the differing parts highlighted in bold. Annotators were asked to determine {\em which of the two sentences accurately describes what can be seen in the image?} In each case, they had to choose between five options: (a) the first, but not the second; (b) the second, but not the first; (c) both of them; (d) neither of the two; and (e) I cannot tell. We collected three annotations for each instance, from three independent workers. 

\paragraph{Annotator selection} We recruited annotators who had an approval rating of 90\% or higher on Amazon Mechanical Turk. We ran an initial, pre-selection study with 10 batches of 100 instances each, in order to identify annotators who understood the instructions and performed the task adequately. The pre-selection batches were first manually annotated by the authors, and we identified `good' annotators based on the criterion that they preferred the caption to the foil at least 70\% of the time. Based on this, we selected a total of 63 annotators. Annotators were paid \$0.05 per item (i.e. per HIT on Mechanical Turk).

\paragraph{Results}
Table \ref{tab:mturk-results} shows, for each instrument, the number of instances in total, as well as the proportion of instances which we consider {\em valid}, that is, those for which at least two out of three annotators chose the caption, {\em but not the foil}, as the text which accurately describes the image. We also show the number of instances for which annotators unanimously (3/3) chose the caption.

\begin{figure*}[t]
    \centering
    \includegraphics[width=\linewidth]{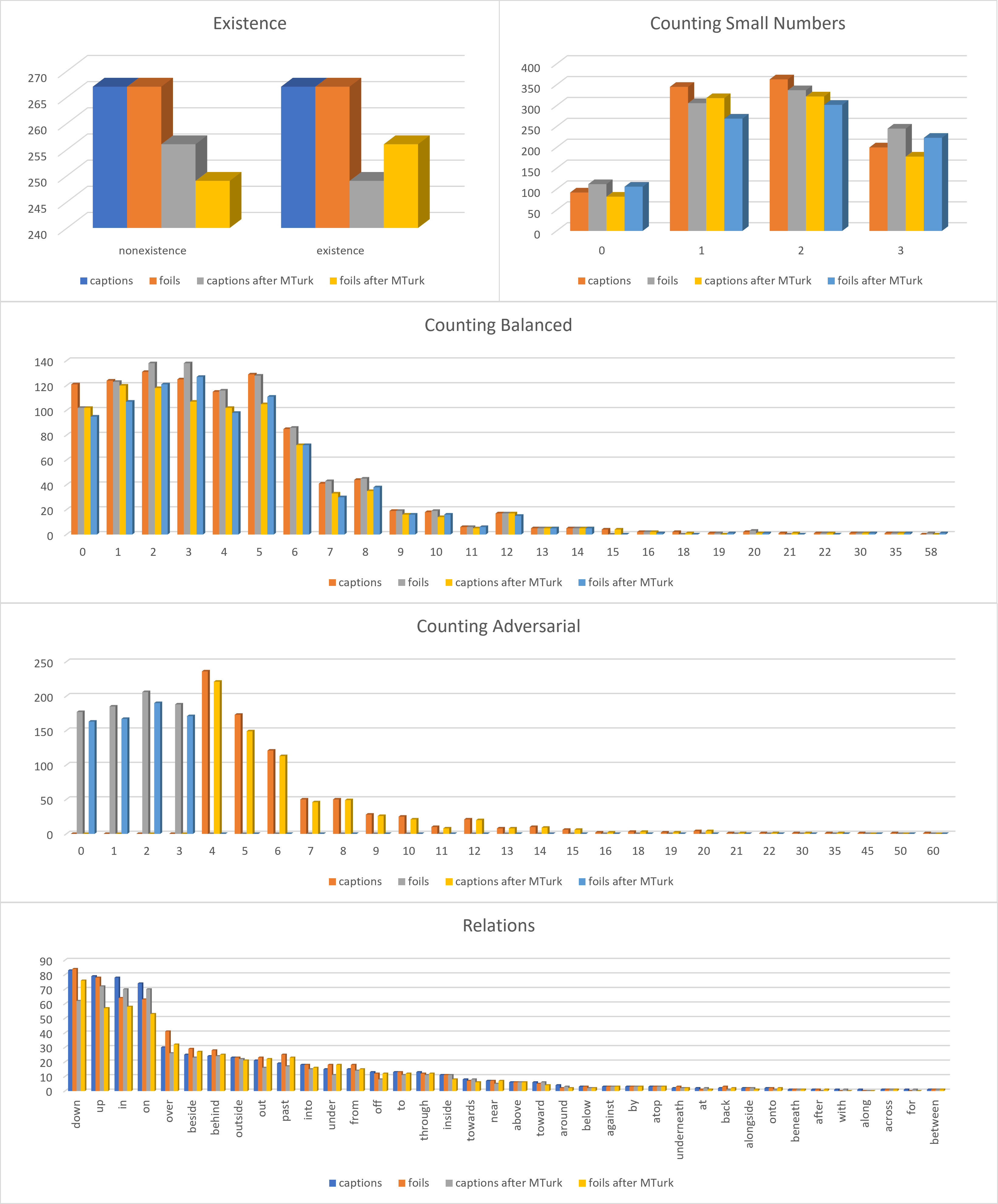}
    \caption{Word frequency distributions for captions and foils before and after the manual validation for existence, counting and relations.}
    \label{fig:foil-caption-distr-part1}
\end{figure*}

\begin{figure*}[t]
    \centering
    \includegraphics[width=\linewidth]{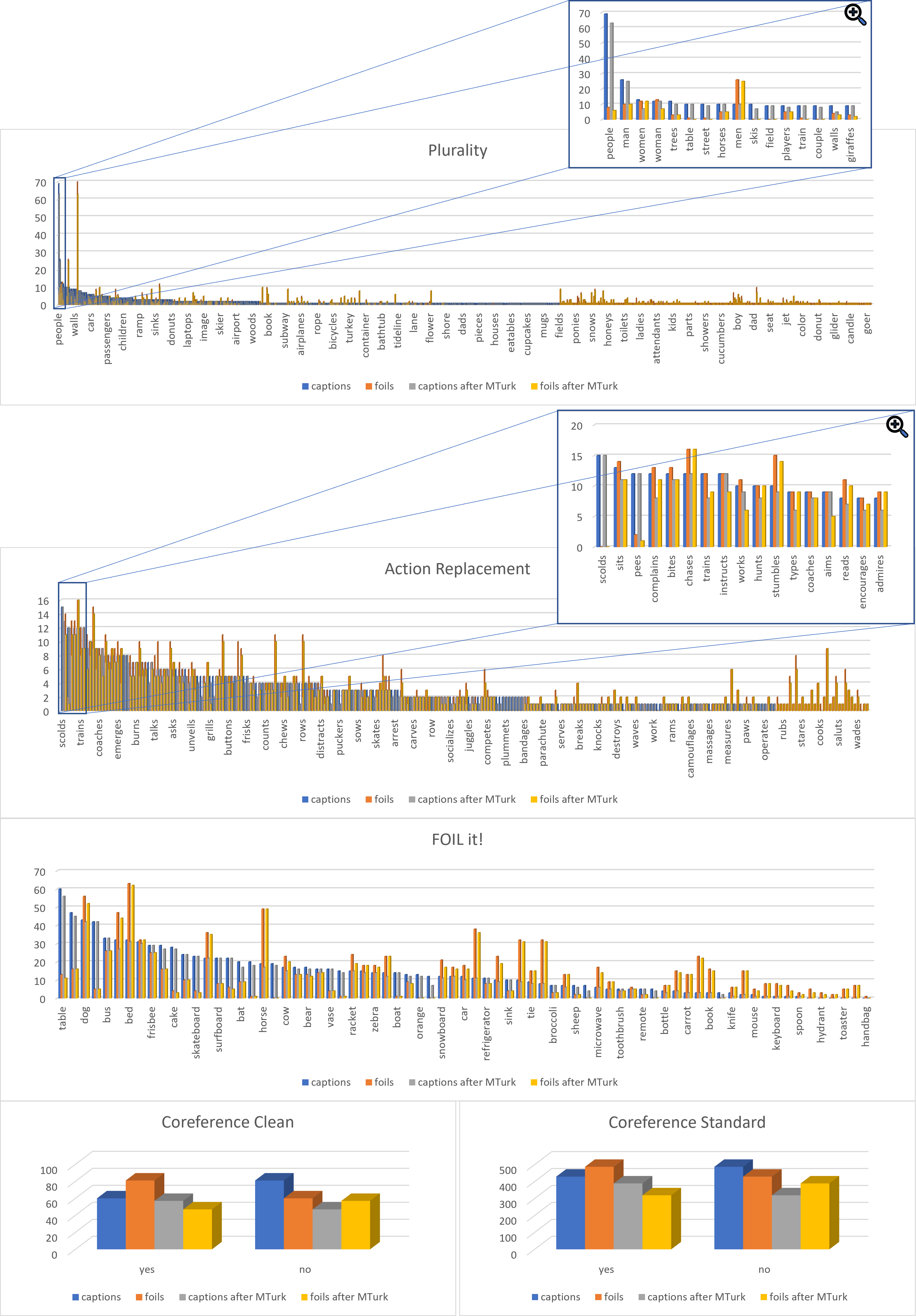}
    \caption{Word frequency distributions for captions and foils before and after the manual validation for plurality, action replacement and FOIL it. The actant swap instrument is not visualised here: By construction, actant swap cannot suffer from distributional bias since caption and foil contain the same words up to a \emph{permutation}.}
    \label{fig:foil-caption-distr-part2}
\end{figure*}


\paragraph{Annotator agreement} As shown in Table \ref{tab:mturk-results}, the proportion of {\em valid} instances in each instrument was high, ranging from 73.8\% to 94.6\%, with most instruments having annotators choose the caption well over 80\% of the time. The table also shows two inter-annotator agreement statistics, both computed using Krippendorff's $\alpha$: over all the data in a given instrument, and over the valid subset only. On the valid subset, agreement is higher, and ranges from 0.3 to 0.6 (mean = 0.42; sd=0.12). There is a significant positive correlation between the percentage of valid instances per instrument and the $\alpha$ value (Spearman's $\rho=0.75; p< .05$).  The low to medium agreement suggested by the $\alpha$ range is due to two factors: first, the statistic is computed over the entire pool of annotators, of whom there were significant diversions in the amount of annotations they computed (e.g. some workers annotated fewer than 5 HITs); furthermore, the agreement is computed over 5 categories (see above). Given these factors, the inter-annotator agreement results should be treated with caution, and are not straightforwardly interpretable as an index of human performance on \dataset{} - in particular, the validation task (with 5 categories) was framed differently from the benchmark (which is binary).

\paragraph{Bias check}
While measures were taken to control for distributional bias between captions and foils in the different pieces of \dataset{} (cf. \S \ref{sec:dist-bias}), it is possible that sub-sampling after manual validation could reintroduce such biases. To check that this is not the case, we compare the \textit{word frequency distributions between captions and foils} in the original pieces, and the word frequency distribution of the manually validated set. We report the Jensen-Shannon divergence and the number of words that differ between caption and foil in Table \ref{tab:mturk-results}. The foil-caption word frequency distributions can be inspected in Figures \ref{fig:foil-caption-distr-part1} and \ref{fig:foil-caption-distr-part2}. The Jensen-Shannon (JS) divergence is defined as:

$$ JS(f \parallel c) = \sqrt{\frac{KL(f \parallel m) + KL(c \parallel m)}{2}} $$

\noindent
where $f$ is the normalized word frequency for foils, $c$ the normalized word frequency for captions, $m$ is the point-wise mean of $f$ and $c$, and $KL$ is the Kullback-Leibler divergence. 

As Table \ref{tab:mturk-results} shows, the JS-divergence between caption and foil distributions remains the same, or changes only marginally (compare columns {\em JS-div} and {\em Js-div valid}, where {\em \#Lexical Items} indicates the number of lexical/phrasal categories in the relevant distributions). This indicates that no significant bias was introduced as a result of subsampling after manual validation.



\begin{table*}[t]
    \small
    \centering
    
    \begin{tabular}{cp{.3\linewidth}b{.25\linewidth}b{.25\linewidth}}
      \toprule
      \bf piece & \bf image & \bf caption (\gr{blue}) & \bf foil (\red{orange}) \\
      \midrule
      \multirow{5}{*}{existence} & \includegraphics[width=\linewidth]{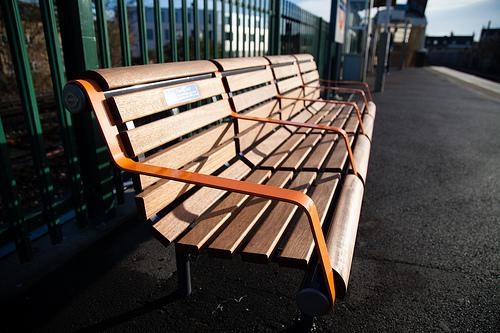}
      & There are \gr{no} people in the picture. & There are people in the picture.\\
      \cmidrule{2-4}
      & \includegraphics[width=\linewidth]{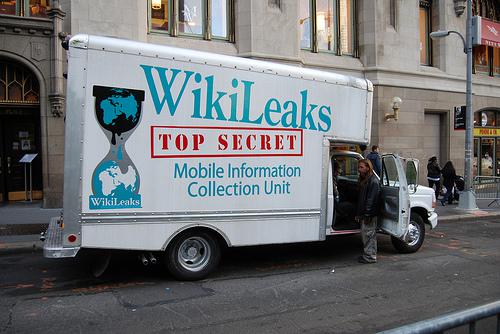}
      & There is a truck pictured. & There is \red{no} truck pictured. \\
      \cmidrule{2-4}
      & \includegraphics[width=\linewidth]{imgs/data-examples/v7w_61601.jpg}
      & There are \gr{no} clouds in the sky. & There are clouds in the sky. \\
      \cmidrule{2-4}
      & \includegraphics[width=\linewidth]{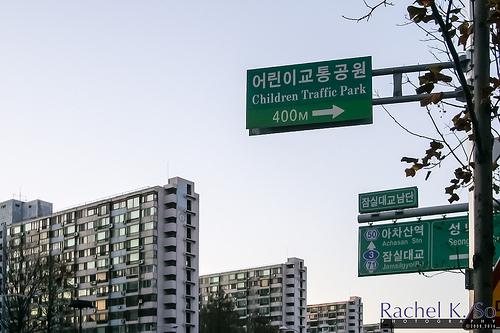}
      & There are \gr{no} people riding on elephants. & There are people riding on elephants. \\
      \cmidrule{2-4}
        & \includegraphics[width=\linewidth]{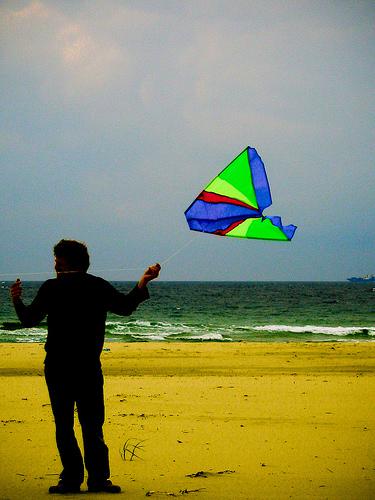}
      & There is a kite. & There is \red{no} kite. \\
      \bottomrule
    \end{tabular}
    \caption{Randomly selected data examples for existence.}
    \label{tab:existence-examples}
\end{table*}

\begin{table*}[t]
    \small
    \centering
    
    \begin{tabular}{cp{.3\linewidth}b{.25\linewidth}b{.25\linewidth}}
      \toprule
    \bf piece & \bf image & \bf caption (\gr{blue}) & \bf foil (\red{orange}) \\
    \midrule
    \multirow{5}{*}{plurality} & \includegraphics[width=\linewidth]{imgs/data-examples/COCO_val2014_000000291619.jpg}
      & Two young men playing frisbee at night on \gr{exactly one sports field}. & Two young men playing frisbee at night on \red{a number of sports fields}. \\
      \cmidrule{2-4}
      & \includegraphics[width=\linewidth]{imgs/data-examples/COCO_val2014_000000455716.jpg}
      & \gr{Exactly one row} of motorcycles parked together on a grass yard area with a house in the background. & \red{A number of rows} of motorcycles parked together on a grass yard area with a house in the background. \\
      \cmidrule{2-4}
      & \includegraphics[width=\linewidth]{imgs/data-examples/COCO_val2014_000000263425.jpg}
      & Two men are looking inside of \gr{a single giant barbecue}. & Two men are looking inside of \red{a number of giant barbecues}. \\
      \cmidrule{2-4}
      & \includegraphics[width=\linewidth]{imgs/data-examples/COCO_val2014_000000457884.jpg}
      & \gr{Some children} are playing baseball outside in a field. & \red{A single child} is playing baseball outside in a field. \\
      \cmidrule{2-4}
    & \includegraphics[width=\linewidth]{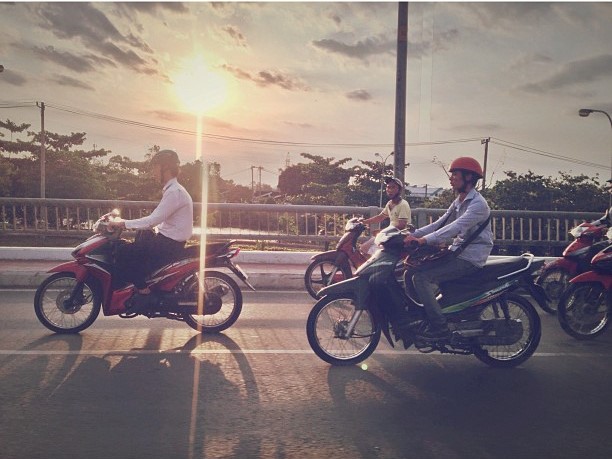}
      & \gr{A number of people} riding some motorbikes on the road. & \red{A single person} riding some motorbikes on the road. \\
      \bottomrule
    \end{tabular}
    \caption{Randomly selected data examples for plurality.}
    \label{tab:plurality-examples}
\end{table*}

\begin{table*}[t]
    \small
    \centering
    
    \begin{tabular}{cp{.3\linewidth}b{.25\linewidth}b{.25\linewidth}}
      \toprule
    \bf piece & \bf image & \bf caption (\gr{blue}) & \bf foil (\red{orange}) \\
    \midrule
    \multirow{5}{*}{counting} & \includegraphics[width=\linewidth]{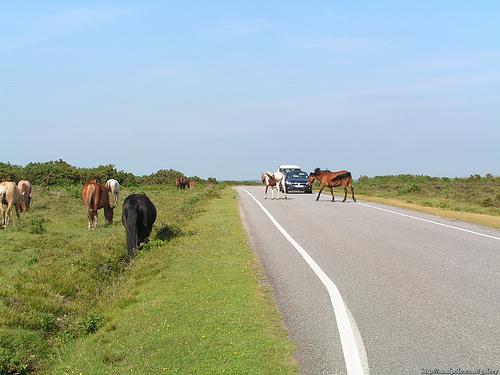}
      & There are exactly \gr{8} horses. & There are exactly \red{5} horses. \\
      \cmidrule{2-4}
      & \includegraphics[width=\linewidth]{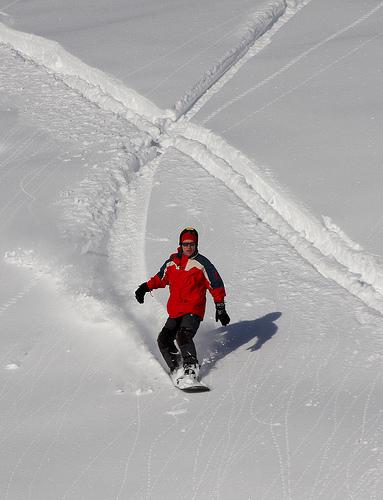}
      & There is exactly \gr{1} person snowboarding. & There are exactly \red{4} people snowboarding. \\
      \cmidrule{2-4}
      & \includegraphics[width=\linewidth]{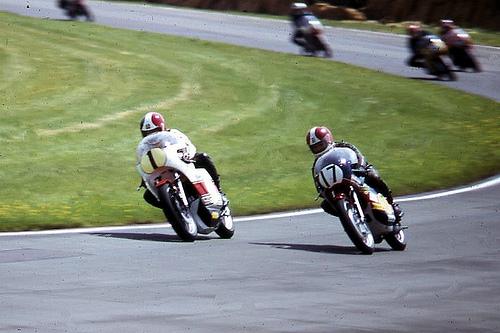}
      & There are exactly \gr{6} motorcycles in this photo altogether. & There are exactly \red{7} motorcycles in this photo altogether. \\
      \cmidrule{2-4}
      & \includegraphics[width=\linewidth]{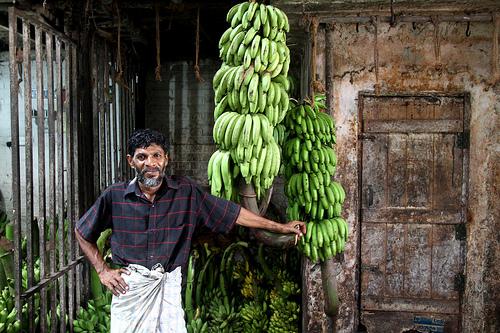}
      & There are exactly \gr{2} banana stalks. & There are exactly \red{4} banana stalks. \\
      \cmidrule{2-4}
    & \includegraphics[width=\linewidth]{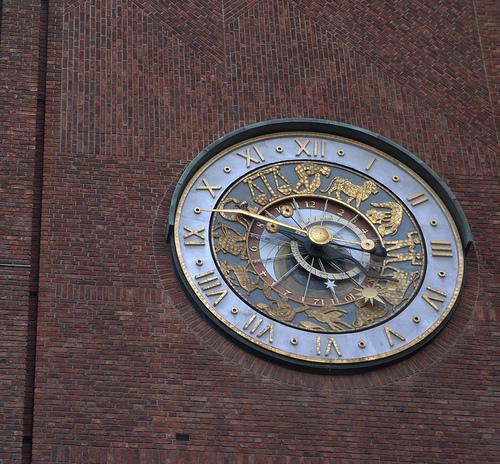}
      & There are exactly \gr{12} roman numerals on the clock. & There are exactly \red{9} roman numerals on the clock. \\
      \bottomrule
    \end{tabular}
    \caption{Randomly selected data examples for counting.}
    \label{tab:counting-examples}
\end{table*}

\begin{table*}[t]
    \small
    \centering
    
    \begin{tabular}{cp{.3\linewidth}b{.25\linewidth}b{.25\linewidth}}
      \toprule
    \bf piece & \bf image & \bf caption (\gr{blue}) & \bf foil (\red{orange}) \\
    \midrule
    \multirow{5}{*}{relations} & \includegraphics[width=\linewidth]{imgs/data-examples/COCO_val2014_000000110638.jpg}
      & A baby elephant is walking \gr{under} a larger elephant. & A baby elephant is walking \red{on} a larger elephant. \\
      \cmidrule{2-4}
      & \includegraphics[width=\linewidth]{imgs/data-examples/COCO_val2014_000000378873.jpg}
      & Fruits and vegetables are being sold \gr{in} a market. & Fruits and vegetables are being sold \red{outside} a market. \\
      \cmidrule{2-4}
      & \includegraphics[width=\linewidth]{imgs/data-examples/COCO_val2014_000000199977.jpg}
      & An airplane is letting \gr{off} white smoke against a blue sky. & An airplane is letting \red{in} white smoke against a blue sky. \\
      \cmidrule{2-4}
      & \includegraphics[width=\linewidth]{imgs/data-examples/COCO_val2014_000000545958.jpg}
      & A cow stands on a sidewalk \gr{outside} a building. & A cow stands on a sidewalk \red{in} a building. \\
      \cmidrule{2-4}
    & \includegraphics[width=\linewidth]{imgs/data-examples/COCO_val2014_000000284296.jpg}
      & Three giraffes banding \gr{down} to drink water with trees in the background. & Three giraffes banding \red{up} to drink water with trees in the background. \\
      \bottomrule
    \end{tabular}
    \caption{Randomly selected data examples for relations.}
    \label{tab:relations-examples}
\end{table*}

\begin{table*}[t]
    \small
    \centering
    
    \begin{tabular}{cp{.3\linewidth}b{.25\linewidth}b{.25\linewidth}}
      \toprule
    \bf piece & \bf image & \bf caption (\gr{blue}) & \bf foil (\red{orange}) \\
    \midrule
    \multirow{5}{*}{actions} & \includegraphics[width=\linewidth]{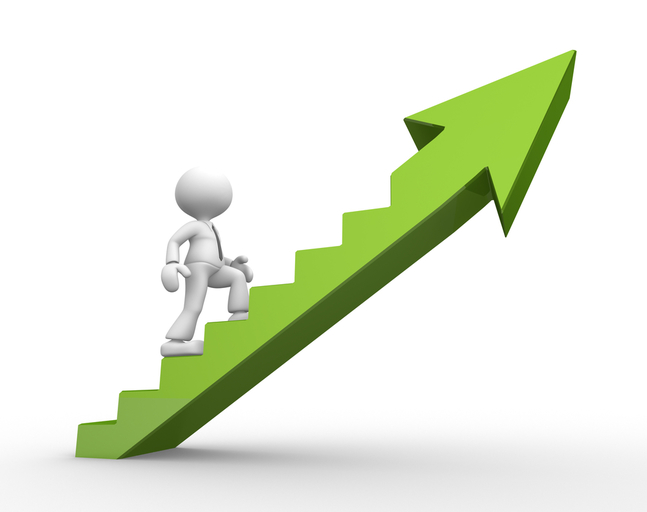}
      & A figure \gr{climbs} the stairs. & A figure \red{descends} the stairs. \\
      \cmidrule{2-4}
      & \includegraphics[width=\linewidth]{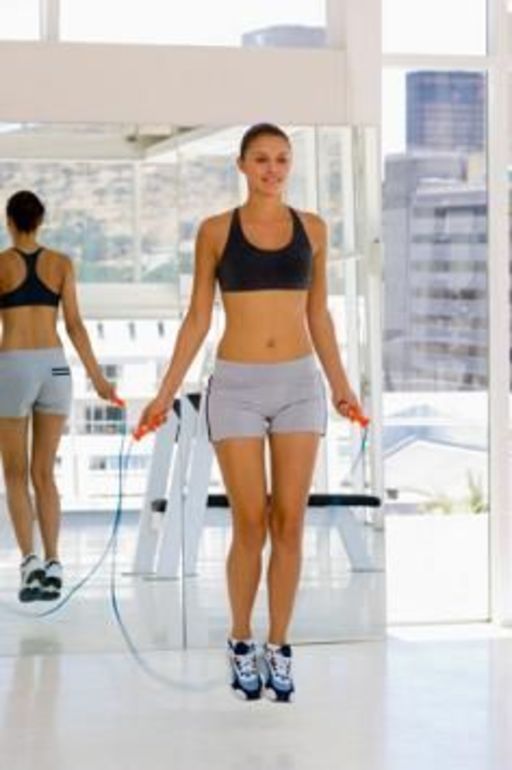}
      & A woman \gr{skips} a jump rope. & A woman \red{releases} a jump rope.\\
      \cmidrule{2-4}
      & \includegraphics[width=\linewidth]{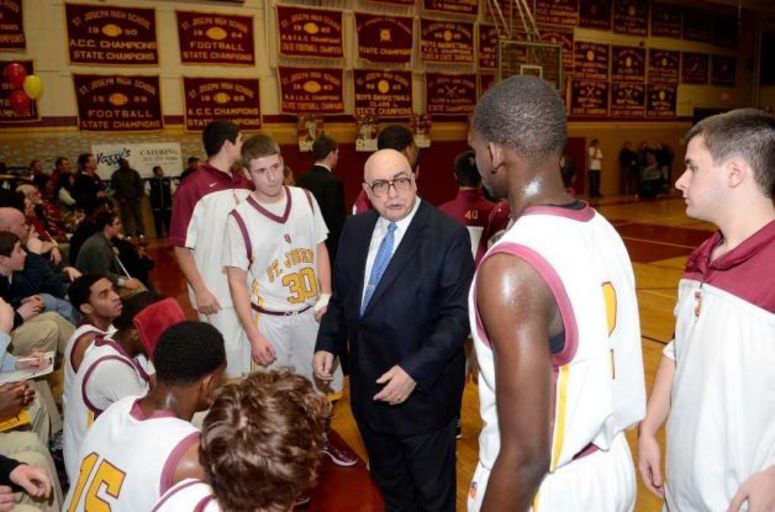}
      & An old man \gr{coaches} people. & An old man \red{bothers} people. \\
      \cmidrule{2-4}
      & \includegraphics[width=\linewidth]{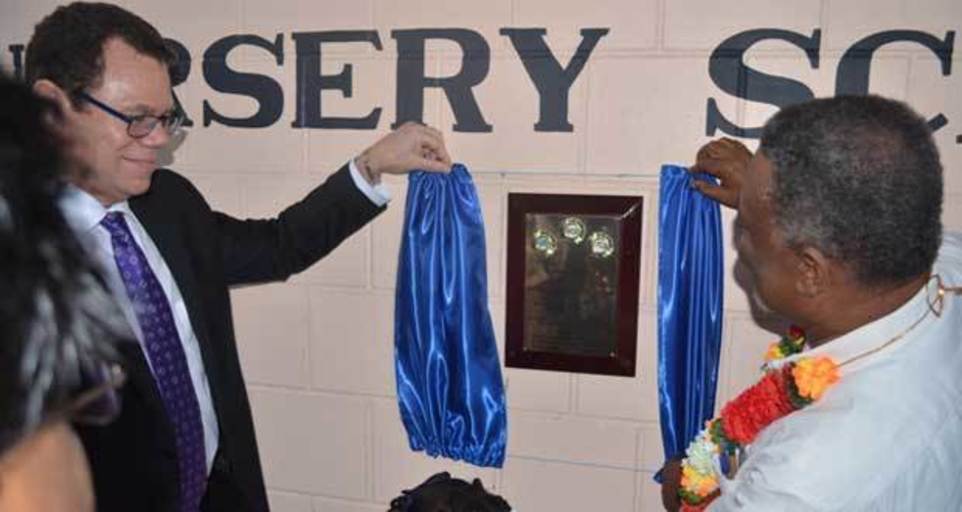}
      & \gr{The people} unveil \gr{the prize}. & \red{A prize} unveils \red{people}. \\
      \cmidrule{2-4}
    & \includegraphics[width=\linewidth]{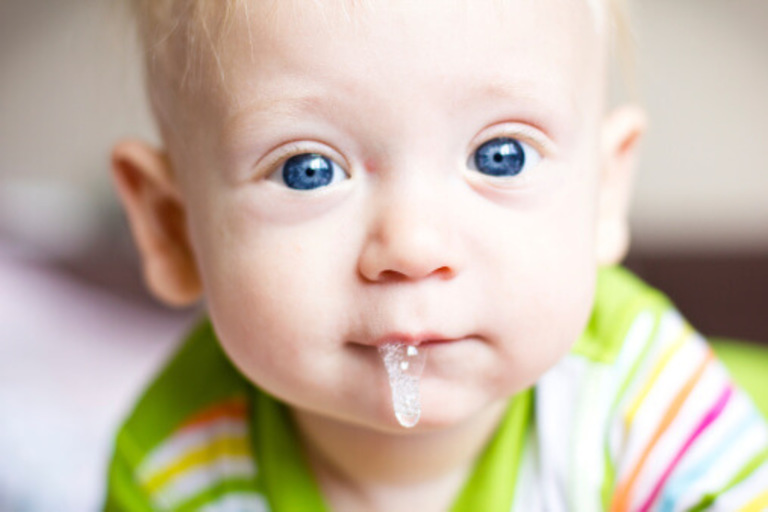}
      & \gr{A baby drools} over clothing. & \red{A clothing} drools over \red{the baby}. \\
      \bottomrule
    \end{tabular}
    \caption{Randomly selected data examples for actions.}
    \label{tab:actions-examples}
\end{table*}

\begin{table*}[t]
    \small
    \centering
    
    \begin{tabular}{cp{.28\linewidth}b{.25\linewidth}b{.25\linewidth}}
      \toprule
    \bf piece & \bf image & \bf caption (\gr{blue}) & \bf foil (\red{orange}) \\
    \midrule
    \multirow{5}{*}{coreference} & \includegraphics[width=.75\linewidth]{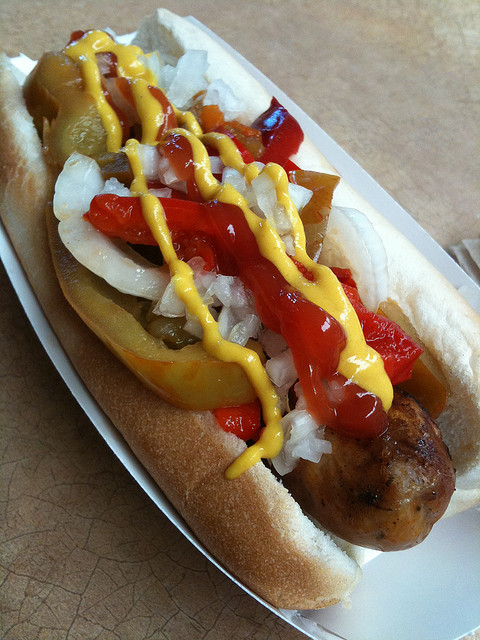}
      & A close up of a hot dog with onions. Is it a big hot dog? \gr{Yes}. & A close up of a hot dog with onions. Is it a big hot dog? \red{No}. \\
      \cmidrule{2-4}
      & \includegraphics[width=\linewidth]{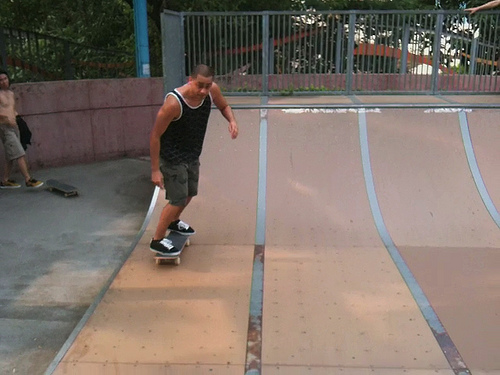}
      & A skateboarding man is on a half pipe. Does he wear a helmet? \gr{No}. & A skateboarding man is on a half pipe. Does he wear a helmet? \red{Yes}. \\
      \cmidrule{2-4}
      & \includegraphics[width=\linewidth]{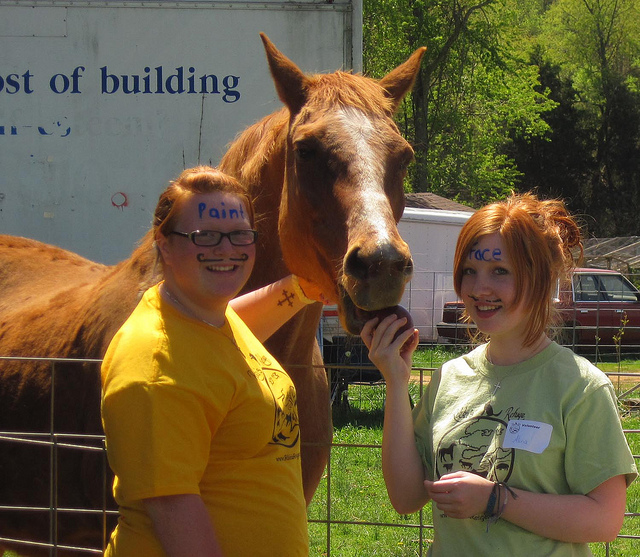}
      & 2 women who have painted on mustaches petting a horse. Are they wearing hats? \gr{No}. & 2 women who have painted on mustaches petting a horse. Are they wearing hats? \red{Yes}. \\
      \cmidrule{2-4}
      & \includegraphics[width=.75\linewidth]{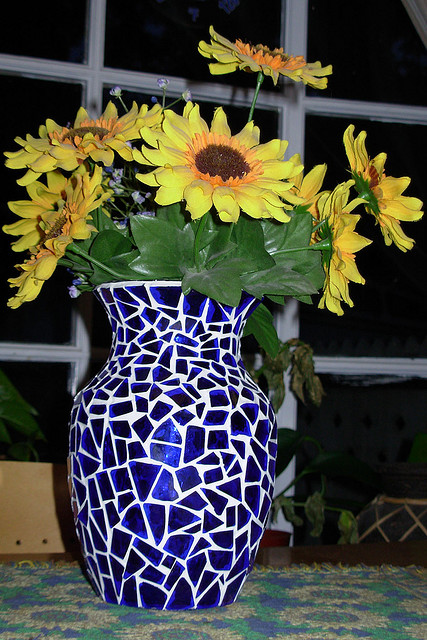}
      & Yellow sunflowers are in a blue and white giraffe styled vase. Is it inside? \gr{Yes}. & Yellow sunflowers are in a blue and white giraffe styled vase. Is it inside? \red{No}. \\
      \cmidrule{2-4}
    & \includegraphics[width=.75\linewidth]{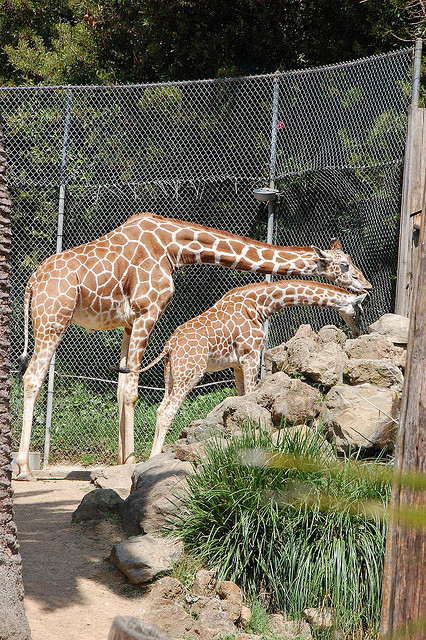}
      & An adult giraffe and a child giraffe standing near a fence. Does this look like zoo? \gr{Yes}. & An adult giraffe and a child giraffe standing near a fence. Does this look like zoo? \red{No}. \\
      \bottomrule
    \end{tabular}
    \caption{Randomly selected data examples for coreference.}
    \label{tab:coreference-examples}
\end{table*}

\end{document}